\documentclass[final,5p,times,twocolumn]{elsarticle}

\usepackage{xcolor}  

\usepackage{amsmath,amssymb,amsfonts}
\usepackage{algorithmic}
\usepackage{array}
\usepackage{graphicx}
\usepackage{url}
\usepackage{enumitem}
\usepackage{arydshln}
\usepackage{multirow}
\usepackage{booktabs}
\usepackage{natbib}
\usepackage{hyperref}
\usepackage[table]{xcolor}

\journal{ISPRS Journal of Photogrammetry and Remote Sensing}

\begin{document}

\begin{frontmatter}



\title{MultiLevel Joint Learning with Spatial and Frequency Domain Enhancement for Cross-View Geo-Localization}



\author{Hongying Zhang\corref{cor1}}
\ead{carole\_zhang@vip.163.com}

\author{ShuaiShuai Ma}

\cortext[cor1]{Corresponding author}

\address{School of Electronic Information and Automation, Civil Aviation University of China, Tianjin, 300300, China}
\begin{abstract}
Cross-view geo-localization (CVGL) aims to establish spatial correspondences between images captured from significantly different viewpoints and constitutes a fundamental technique for visual localization in GNSS-denied environments. Nevertheless, CVGL remains challenging due to severe geometric asymmetry, texture inconsistency across imaging domains, and the progressive degradation of discriminative local information. Existing methods predominantly rely on spatial domain feature alignment, which is inherently sensitive to large scale viewpoint variations and local disturbances.  To alleviate these limitations, this paper proposes the Spatial and Frequency Domain Enhancement Network (SFDE), which leverages complementary representations from spatial and frequency domains. SFDE adopts a three branch parallel architecture to model global semantic context, local geometric structure, and statistical stability in the frequency domain, respectively, thereby characterizing consistency across domains from the perspectives of scene topology, multiscale structural patterns, and frequency invariance. The resulting complementary features are jointly optimized in a unified embedding space via progressive enhancement and coupled constraints, enabling the learning of cross-view representations with consistency across multiple granularities. Comprehensive experiments show that SFDE achieves competitive performance and in many cases even surpasses state-of-the-art methods, while maintaining a lightweight and computationally efficient design. {Our code is available at {\href{https://github.com/Mashuaishuai669/SFDE}{\textcolor{magenta}{https://github.com/Mashuaishuai669/SFDE}}}}. 
\end{abstract}



\begin{keyword}
Cross-view geo-localization, Image retrieval, Multiscale geometric modeling, Frequency domain enhancement
\end{keyword}

\end{frontmatter}



\section{Introduction}
\label{sec:intro}

Cross-view geo-localization (CVGL) has attracted increasing attention due to its importance in autonomous driving, aerial photography, and autonomous navigation. By learning semantic correspondences between a query image and a geo-referenced image database, CVGL supports accurate estimation of geographic coordinates~\cite{lin2013cross}. Early studies mainly focused on matching ground-view images with satellite images~\cite{workman2015location, liu2019lending}. In recent years, with the widespread deployment of unmanned aerial vehicles (UAVs) for urban monitoring and disaster response, research has gradually shifted toward matching UAV images with satellite images for localization under GNSS-denied conditions~\cite{zheng2020university,chen2025efficient,YE2024306}. Nevertheless, pronounced appearance discrepancies and spatial misalignment between UAV and satellite views continue to pose substantial challenges for CVGL~\cite{WANG2025362,chen2025without}.

Early CVGL methods mainly relied on handcrafted feature extraction techniques~\cite{lowe2004distinctive, dalal2005histograms}, using manually designed descriptors to capture representative visual cues from satellite and ground-view images for cross-view matching and localization. However, handcrafted features show limited robustness under complex conditions such as viewpoint variations, scale discrepancies, and appearance inconsistencies, leading to constrained localization accuracy.

With the advent of deep learning, CVGL research has achieved substantial progress~\cite{long2017accurate}. Early deep methods adopted pretrained convolutional networks to extract global descriptors and learned cross-domain embedding spaces through metric learning~\cite{chopra2005learning, arandjelovic2016netvlad}. However, image-level representations proved insufficient for handling local geometric variations~\cite{shi2020optimal}. Subsequent studies introduced region weighting, spatial attention, and alignment mechanisms based on spatial partitioning to enhance local feature learning~\cite{lin2022joint}. Recent advances have explored multiscale feature fusion~\cite{TANG202565, LI202588}, cross-domain geometric transformations~\cite{zhang2023cross, zhang2024geodtr+}, and contrastive learning strategies~\cite{wu2024camp}. These approaches have demonstrated consistent improvements on standard benchmarks~\cite{zheng2020university, zhu2023sues}. Despite these advances, existing methods still experience pronounced performance degradation under extreme viewpoint changes and challenging cross-domain generalization settings~\cite{wang2024multiple}.

The first fundamental challenge stems from the pronounced sensitivity of spatial domain feature learning to cross-domain discrepancies. Most existing methods learn feature correlations in the spatial domain through convolutional receptive fields or local attention windows~\cite{liu2022convnet, liu2021swin}, implicitly assuming structural stability within local spatial neighborhoods. However, the geometric asymmetry between oblique UAV images and orthorectified satellite projections leads to substantial structural inconsistencies for the same objects across different views. Perspective distortion introduces building facades while deforming rooftop contours~\cite{zhai2017predicting}, occlusions cause local information loss~\cite{shi2020where}, and non-uniform scale variations disrupt spatial metric consistency~\cite{chen2024sdpl}. These geometric misalignments break the neighborhood consistency assumptions underlying convolutional and attention-based spatial representations~\cite{xia2024enhancing}, resulting in pronounced instability of spatial domain features under severe viewpoint variations.

The second challenge lies in the limited and unsystematic exploitation of statistical stability in the frequency domain. The Fourier transform decomposes images into energy distributions across different spatial frequencies, where the amplitude spectrum captures global energy organization and the phase spectrum preserves spatial geometric relationships. Previous studies indicate that under CVGL conditions, low-frequency energy distributions follow stable power-law decay patterns and phase gradients maintain topological invariance, both exhibiting stronger statistical consistency than spatial domain textures~\cite{yin2019fourier}. However, most existing approaches incorporate frequency domain cues only through shallow operations such as band decomposition or spectral enhancement~\cite{WANG2025362}, failing to fully exploit the complementary roles of amplitude and phase information, lacking adaptive emphasis on discriminative frequency components~\cite{zeng2025frequency}, and not establishing effective mechanisms to integrate spatial and frequency representations in a unified manner.

To address these challenges, we propose the Spatial and Frequency Domain Enhancement Network (SFDE), a deep neural network framework that learns spatial domain and frequency domain representations in a coordinated manner. The proposed method extracts, enhances, and fuses features from both domains through parallel branches to support collaborative learning of image details, geometric structures, and globally consistent distributions. Specifically, coarse-grained features produced by a ConvNeXt-Tiny backbone~\cite{liu2022convnet} are forwarded to three dedicated branches: the Global Semantic Consistency Branch (GSCB), the Local Geometric Sensitivity Branch (LGSB), and the Frequency Stability Alignment Branch (FSAB). The GSCB captures macroscopic structural cues through global pooling, while the LGSB learns geometric responses ranging from fine-grained edges to midlevel contours by combining multidilation convolutions~\cite{yu2016dilated}, attention interactions, and a learnable spatial pyramid~\cite{he2015spatial, shen2023mccg, zhao2024transfg}. The FSAB separates amplitude and phase components and applies a three-layer adaptive frequency reweighting strategy to capture statistical stability in the frequency domain~\cite{stuchi2024frequency}.

The main contributions of this work are summarized as follows:
\begin{enumerate}
    \item We present a multilevel joint learning framework that treats CVGL as a unified optimization task across three complementary structural dimensions.
    \item We develop the LGSB based on multiscale dilated convolutions and a learnable pyramid structure, which captures spatial relationships ranging from local textures to mid range geometric configurations.
    \item We introduce the FSAB that leverages spectral statistical stability by jointly exploiting amplitude and phase information with adaptive frequency regulation.
    \item Extensive experiments demonstrate that SFDE delivers competitive performance and even outperforms existing methods in several scenarios. Notably, it maintains a lightweight architecture that achieves an effective balance between computational efficiency and localization accuracy.
\end{enumerate}

\section{RELATED WORK}
\label{sec:related}

\subsection{Cross-View Geo-Localization}
CVGL seeks to establish correspondences between images acquired from different imaging platforms and viewpoints~\cite{lin2013cross}. The primary difficulty of this task arises from pronounced viewpoint discrepancies and geometric inconsistencies across imaging modalities, causing identical objects to exhibit substantial appearance variations when observed from different perspectives~\cite{shi2020where}.

Early CVGL methods relied on handcrafted feature descriptors such as SIFT~\cite{lowe2004distinctive} and HOG~\cite{dalal2005histograms} for cross-view matching, yet their performance degraded significantly under large viewpoint changes. With the rapid development of deep learning, convolutional neural networks (CNNs) have emerged as the dominant approach for feature extraction in CVGL~\cite{long2017accurate}. Workman and Jacobs first adopted a CNN pretrained on ImageNet to extract deep features for CVGL~\cite{workman2015location}. Subsequent work further fine-tuned these networks on CVGL datasets, enabling the extraction of shared semantic cues across viewpoints and leading to notable improvements in localization accuracy.

Structural and spatial constraint methods attempt to mitigate viewpoint discrepancies by explicitly exploiting geometric relationships. For instance, Tian et al.~\cite{tian2021uav} introduced topological relationships among multiple buildings within a single area to support geo-localization, while Liu and Li~\cite{liu2019lending} improved spatial consistency by incorporating orientation encoding between ground-view and satellite images. As research advanced, local region based strategies became increasingly important for enhancing the discriminative capability of CVGL matching. Wang et al.~\cite{wang2021each} established cross-view spatial associations at the local region level through blockwise weighting schemes, whereas Dai et al.~\cite{dai2021transformer} adopted Transformer architectures to capture regionwise interactions, thereby improving robustness under complex environmental conditions.

Geometric alignment and multiscale representation methods further seek to mitigate CVGL discrepancies through explicit geometric transformations. Shi et al.~\cite{shi2020optimal} proposed converting satellite images into polar coordinates to reduce viewpoint inconsistencies, though this strategy inevitably introduces geometric distortion and information loss. To alleviate this limitation, Shi et al.~\cite{shi2022accurate} later employed generative adversarial networks to recover the transformed images. In recent years, deep backbone architectures and contrastive learning strategies have been incorporated into CVGL between UAV and satellite images. Shen et al.~\cite{shen2023mccg} adopted ConvNeXt as the feature extraction backbone combined with attention mechanisms to improve localization robustness, while Deuser et al.~\cite{deuser2023sample4geo} leveraged contrastive learning to enhance representation discrimination. More recent work by Chen et al.~\cite{chen2025multilevel} introduced progressive multilevel augmentation with consistency and invariance learning to strengthen cross-domain alignment, yielding strong performance on standard benchmarks~\cite{zhu2023sues, wang2024multiple}.

Despite notable progress in spatial domain feature learning, local alignment strategies, and multiscale representations~\cite{lv2024direction, ge2024multibranch}, most existing approaches still rely on local neighborhood relationships or explicit spatial alignment to enforce cross-view consistency~\cite{zhang2023cross, zhang2024geodtr+}. Owing to the pronounced geometric asymmetry between oblique UAV images and orthorectified satellite projections, perspective distortion, occlusion, and non-uniform scale variations~\cite{chen2024sdpl} introduce substantial structural discrepancies for identical objects across different views. Such discrepancies weaken the reliability of features defined on fixed spatial scales or local neighborhoods under severe viewpoint changes~\cite{xia2024enhancing}. Most existing methods attempt to alleviate these limitations through engineered region-based strategies or attention mechanisms~\cite{hu2022beyond}, yet robustness to geometric perturbations and multiscale spatial misalignment has not been systematically addressed at the level of feature representation design.

To address these limitations, we introduce the LGSB that learns spatial relationships ranging from local textures to mid range geometric configurations by combining multiscale dilated convolutions~\cite{yu2016dilated} with a learnable spatial pyramid structure~\cite{he2015spatial}. This design improves representation stability under pronounced viewpoint variations and geometric perturbations, which provides a more reliable foundation for CVGL.

\subsection{Frequency Domain Alignment}

Representations in the frequency domain decompose images into components at different spatial frequencies through the Fourier transform, which offers a complementary perspective to representations in the spatial domain for visual feature learning. Previous studies have shown that low frequency components primarily capture global structure and energy distribution, whereas high frequency components encode local details such as edges and textures. Building on these observations, frequency domain representations have been widely applied in computer vision tasks and have exhibited distinct advantages for feature extraction across domains and robustness under domain shifts.

In image generation and reconstruction, Gatys et al.~\cite{gatys2016image} showed that the amplitude spectrum primarily reflects texture characteristics while the phase spectrum preserves structural information, highlighting the role of spectral statistics in describing image properties. In studies of model robustness, Yin et al.~\cite{yin2019fourier} reported that deep networks are particularly sensitive to high-frequency perturbations, and that constraining frequency components or attenuating high-frequency noise can improve stability and generalization. In the context of domain adaptation and cross-domain representation learning, Yang et al.~\cite{yang2020fda} further observed that different data domains often share higher consistency in low-frequency statistics, offering an effective direction for mitigating domain shift.

These studies suggest that frequency features maintain stronger statistical stability across different domains and data distributions. This property is particularly relevant to CVGL: although UAV and satellite images differ substantially in viewpoint and geometric configuration in the spatial domain, their frequency domain representations, especially low frequency energy distributions, can remain relatively consistent across viewpoints and scales. Such consistency provides an additional dimension for improving CVGL alignment and matching.

Only a limited number of studies have examined the role of frequency domain information in CVGL. For example, FENet proposed by Zeng et al.~\cite{zeng2025frequency} improves robustness through frequency enhancement strategies. Nevertheless, existing approaches still face two fundamental limitations. First, frequency domain information is commonly used as an auxiliary enhancement signal rather than being treated as an explicit and independently learnable discriminative representation space. Although the amplitude and phase spectra respectively reflect global energy distribution and spatial structure, their differences in stability and their complementary roles under varying viewpoints have not been sufficiently investigated. Second, different frequency components contribute unequally to CVGL matching, yet most existing methods apply uniform frequency processing strategies and lack data-driven mechanisms for adaptive frequency selection~\cite{stuchi2024frequency}.

To address these limitations, we introduce the FSAB that leverages statistical regularities in the frequency domain by jointly exploiting amplitude and phase information together with an adaptive frequency reweighting strategy. By complementing spatial domain features, this branch supports more reliable CVGL alignment and matching.

\begin{figure*}[t]
  \centering
  \includegraphics[width=7.1in]{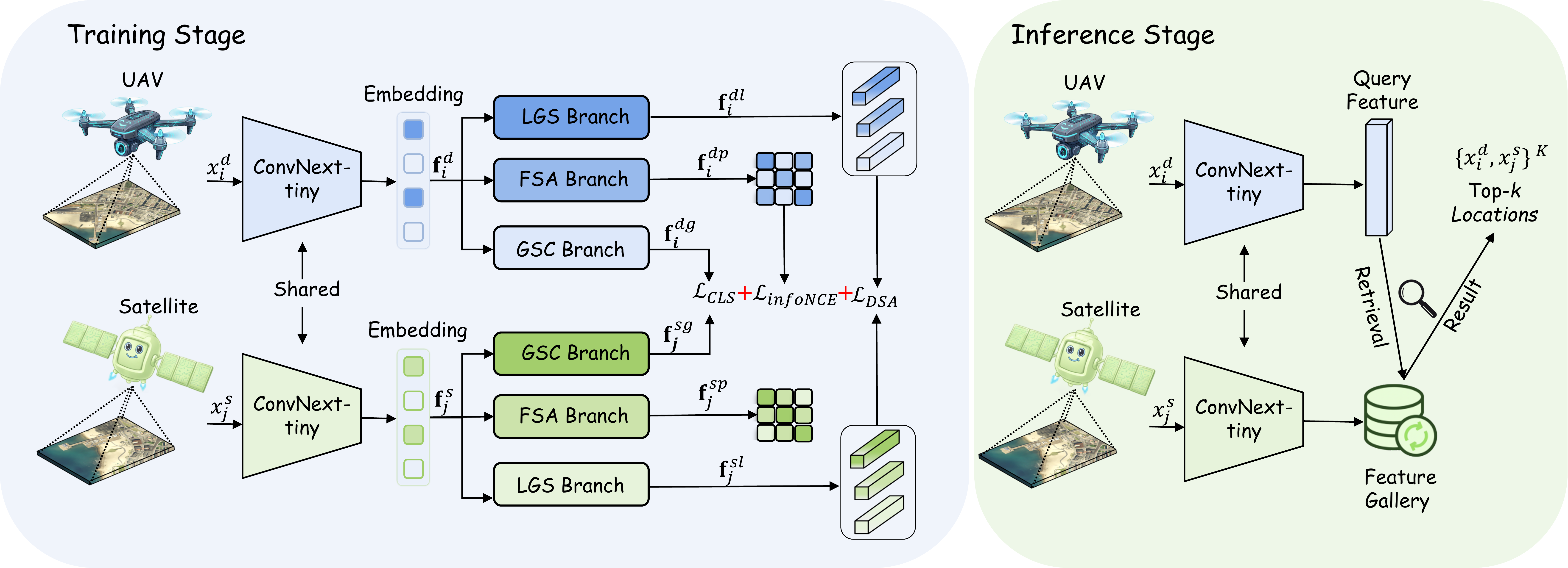}
 \caption{Overall architecture of the proposed SFDE network. The network adopts a three branch parallel design to capture GSCB, LGSB, and FSAB from complementary perspectives. The left part depicts the shared backbone for feature extraction and the subsequent multi branch processing, while the right part illustrates the complete inference workflow.}
  \label{fig1}  
  \end{figure*}
  
\section{Proposed Method}

The overall architecture of the proposed SFDE network is illustrated in Fig.~\ref{fig1}. The proposed CVGL network adopts a three branch parallel architecture to characterize CVGL features from three complementary perspectives: global semantic consistency, local geometric sensitivity, and statistical stability in the frequency domain. The GSCB aggregates global contextual information from local descriptors using an expanded receptive field, retaining discriminative semantics while suppressing fine-grained noise. The LGSB focuses on spatial relationships at multiple scales, capturing geometric cues ranging from local textures to midlevel structural patterns. The FSAB leverages the power-law distribution of spectral energy in natural scenes, preserving low-frequency structural components while emphasizing discriminative high-frequency details. During training, the three branches are supervised using cross-entropy loss, contrastive learning loss, and cross-domain alignment loss, respectively. Through this multigranularity design, the network captures cross-domain consistency at different abstraction levels and maintains robustness under large viewpoint variations and complex imaging conditions.

\textit{Problem Formulation.}
Given a CVGL dataset covering $P$ geographic regions, let $\{x_i^{d}\}_{i=1}^{N}$ represent the set of UAV-view images and $\{x_j^{s}\}_{j=1}^{M}$ represent the set of satellite-view images, where each UAV image $x_i^{d}$ is associated with a corresponding satellite image $x_j^{s}$, forming a positive pair $(x_i^{d}, x_j^{s})$. The goal of CVGL is to learn an embedding function $f_{\theta}:\mathcal{X}\rightarrow \mathbb{R}^{D}$ that maps images into a shared feature space in which the distance between positive pairs is reduced, while the distance between negative pairs is increased. During inference, a query image is projected into the embedding space using the learned representation function $f_{\theta}$ and compared with embeddings of all reference gallery images based on cosine similarity. The Top-$K$ nearest neighbors are then returned as localization candidates.

\subsection{ConvNeXt-Tiny Backbone for Feature Extraction}

In this work, we adopt the lightweight ConvNeXt-Tiny network as the backbone feature extractor. Given a pair of cross-view images $(x_i^{d}, x_j^{s})$, both images are processed by a weight shared backbone network to obtain deep semantic feature representations, which are expressed as
\begin{equation}
f_i^{d} = \mathcal{F}_{\mathrm{backbone}}(x_i^{d}), \quad
f_j^{s} = \mathcal{F}_{\mathrm{backbone}}(x_j^{s}),
\end{equation}
where $\mathcal{F}_{\mathrm{backbone}}(\cdot)$ denotes the ConvNeXt-Tiny feature extractor. The resulting feature maps have the shape $f_i^{d} \in \mathbb{R}^{C \times H \times W}$ and $f_j^{s} \in \mathbb{R}^{C \times H \times W}$, where $C$, $H$, and $W$ represent the channel dimension, height, and width of the feature maps, respectively.

\subsection{Global Semantic Consistency Branch}

The feature maps $f_i^{d}$ and $f_j^{s}$ extracted by the backbone network encode both semantic channel information and spatial distribution characteristics. The channel dimension $C$ aggregates semantic responses across hierarchical levels, while the spatial dimension $H \times W$ retains the relative positions and neighborhood structures of local regions. Consequently, $f_i^{d}$ and $f_j^{s}$ can be interpreted as collections of $H \times W$ local feature units describing textures, edges, and structural patterns within local spatial neighborhoods. However, different geographic regions may exhibit highly similar global structural layouts, causing CVGL matching that relies only on local feature responses to struggle with distinguishing global layout differences and leading to false matches and ambiguity. To alleviate this issue, we introduce the GSCB that constrains CVGL representation learning by establishing stable semantic anchors at a global level. Specifically, taking the UAV feature $f_i^{d}$ as an example, this branch first applies global average pooling over the spatial dimensions to aggregate $f_i^{d}$ into a global descriptor, which is subsequently refined through a diversified embedding classifier (DEC) module $DEC(\cdot)$~\cite{chen2025multilevel} to enhance discriminability. This process is formulated as
\begin{equation}
f_i^{dg}
=
DEC(
\frac{1}{H \times W}
\sum_{h=1}^{H}
\sum_{w=1}^{W}
f_i^{d}
),
\end{equation}
while $f_j^{sg}$ is obtained in the same manner.

\begin{figure}[t]
  \centering
  \includegraphics[width=3.3in]{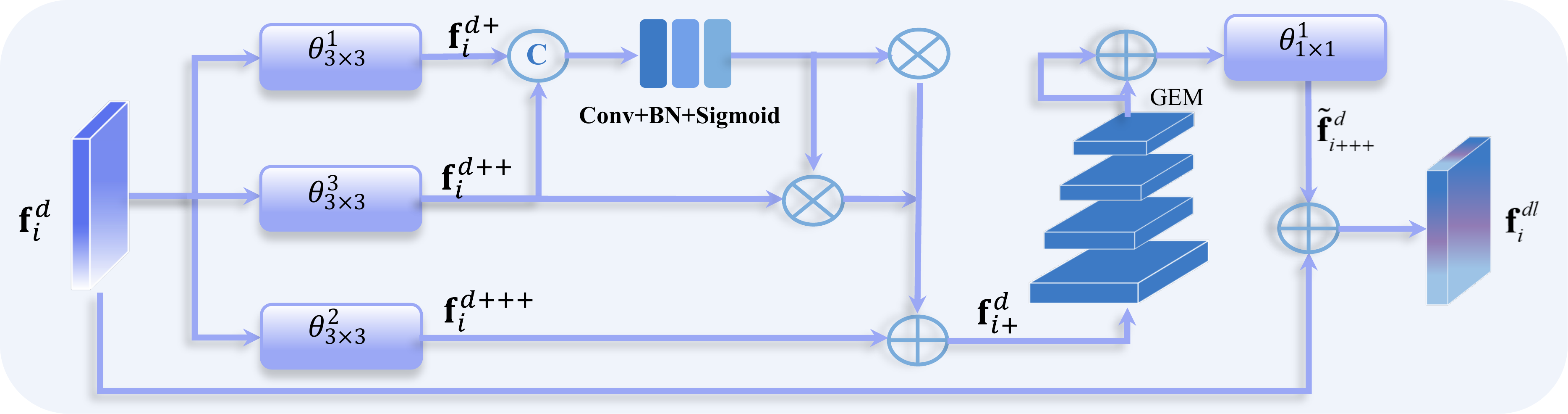}
  \caption{Overview of the LGSB. This branch captures spatial relationships ranging from local textures to mid range geometric configurations via multiscale dilated convolutions, and integrates interactive attention between local and global features with adaptive spatial pyramid pooling to achieve multigranularity geometric-sensitive modeling.
}
  \label{fig2}  
  \end{figure}

\subsection{Local Geometric Sensitivity Branch}

As shown in Fig.~\ref{fig2}, global semantic alignment alone is insufficient for establishing stable local correspondences across viewpoints. The geometric asymmetry between oblique UAV imaging and orthorectified satellite projection~\cite{tian2021uav} introduces perspective distortion, scale shifts, and spatial displacement that require explicit modeling. We therefore introduce the LGSB, which operates on backbone-extracted feature maps that retain spatial structure, namely $f_i^{d} \in \mathbb{R}^{C \times H \times W}$ and $f_j^{s} \in \mathbb{R}^{C \times H \times W}$. By preserving spatial dimensionality, this branch provides geometric cues for subsequent cross-domain matching and improves robustness to scale variation and spatial misalignment.

Taking the UAV feature $f_i^{d} \in \mathbb{R}^{C \times H \times W}$ as an example, three parallel $3 \times 3$ convolutional operations are applied to capture geometric information at multiple spatial scales. These convolutions adopt dilation rates of 1, 2, and 3, denoted as $\theta^{1}_{3 \times 3}$, $\theta^{2}_{3 \times 3}$, and $\theta^{3}_{3 \times 3}$, respectively. Increasing the dilation rate progressively expands the receptive field, which allows local texture responses to transition toward larger-range structural cues. At the same time, the channel dimension is reduced to one quarter of the original size, yielding more compact feature representations that retain multiscale geometric sensitivity. The resulting features are denoted as $f_i^{d+}$, $f_i^{d++}$, and $f_i^{d+++}$. Except for the parallel dilated convolution branches introduced here, all other convolutional layers use a default dilation rate of $1$ without additional dilation. This process can be formulated as
\begin{equation} \begin{aligned} f_i^{d+} &= \theta^{1}_{3 \times 3}(f_i^{d}),\\ f_i^{d++} &= \theta^{2}_{3 \times 3}(f_i^{d}),\\ f_i^{d+++} &= \theta^{3}_{3 \times 3}(f_i^{d}). \end{aligned} \end{equation}
where $f_i^{d+} \in \mathbb{R}^{{C/4}\times H \times W}$, $f_i^{d++} \in \mathbb{R}^{{C/4}\times H \times W}$, and $f_i^{d+++} \in \mathbb{R}^{{C/4}\times H \times W}$.

In addition, We introduce an interaction attention mechanism to enhance the discriminative capacity of multiscale features by fusing local and global information. Specifically, the fine grained local feature $f_i^{d+}$ produced by the smallest receptive-field branch and the coarse grained global feature $f_i^{d+++}$ obtained from the largest receptive-field branch are concatenated along the channel dimension to form a complementary representation that combines local detail and global context. A $1\times1$ convolution $\theta_{1\times1}$ is then applied to capture inter-channel relationships, followed by batch normalization $\mathcal{N}(\cdot)$ and a Sigmoid activation $\delta(\cdot)$ to generate the attention weight map $\omega_1$, which can be formulated as
\begin{equation}
\omega_1
=
\delta\!\left(
\mathcal{N}\!\left(
\theta_{1\times1}\!\left(
\mathrm{cat}\!\left(f_i^{d+},\, f_i^{d+++}\right)
\right)
\right)
\right).
\end{equation}

The resulting attention weights are used to fuse the enhanced local and global features with the intermediate scale feature $f_i^{d++}$ through weighted averaging, producing a multiscale representation $f_{i+}^{d} \in \mathbb{R}^{{C/4}\times H \times W}$, which is expressed as
\begin{equation}
f_{i+}^{d}
=
\frac{1}{3}
\left(
\omega_1 f_i^{d+}
+
f_i^{d++}
+
(1-\omega_1) f_i^{d+++}
\right).
\end{equation}

To further improve the ability of $f_{i+}^{d} \in \mathbb{R}^{{C/4}\times H \times W}$ to capture contextual information across different spatial resolutions, we adopt an adaptive spatial pyramid strategy combined with generalized mean pooling. Given the fused feature $f_{i+}^{d} \in \mathbb{R}^{{C/4}\times H \times W}$, a spatial pyramid with four scales $s \in \{1,2,3,4\}$ is constructed. For clarity, the case of $s=1$ is described. Adaptive average pooling $\Lambda_{1}(\cdot)$ is first applied to compress the spatial dimensions of $f_{i+}^{d} \in \mathbb{R}^{{C/4}\times H \times W}$ into a fixed grid of size $\mathbb{R}^{{C/4}\times 1 \times 1}$. The pooled feature is then processed by a $1\times1$ convolution $\theta_{1\times1}$, batch normalization $\mathcal{N}(\cdot)$, and a nonlinear activation $F_{\mathrm{ReLU}}$ to reorganize channel semantics, after which upsampling $U(\cdot)$ restores the feature to $\mathbb{R}^{{C/4}\times H \times W}$, yielding $f_{i+}^{d1}$. To avoid manually specifying the contributions of different scales and to improve scale adaptivity, learnable scale coefficients $\alpha \in \mathbb{R}^{|S|}$ are introduced and normalized by Softmax to obtain weights $\omega_s$, which are used to recalibrate $f_{i+}^{d1}$. This process can be formulated as
\begin{equation}
f_{i+}^{d1}
=
U\!\left(
F_{\mathrm{ReLU}}\!\left(
\mathcal{N}\!\left(
\theta_{1\times1}\!\left(
\Lambda_{1}\!\left(f_{i+}^{d}\right)
\right)
\right)
\right)
\right),
\end{equation}
\begin{equation}
\tilde{f}_{i+}^{d1}
=
\omega_s f_{i+}^{d1},
\end{equation}
where $\tilde{f}_{i+}^{d1} \in \mathbb{R}^{{C/4}\times H \times W}$, and
$\tilde{f}_{i+}^{d2}$, $\tilde{f}_{i+}^{d3}$, and $\tilde{f}_{i+}^{d4}$ are obtained in the same manner.

Subsequently, features from all pyramid scales are concatenated along the channel
dimension, producing a feature map with $C$ channels. A $1\times1$ convolution
$\theta_{1\times1}$ is then applied to compress the concatenated feature to
$C/4$ channels, resulting in a pyramid-enhanced feature
$\tilde{f}_{i++}^{d} \in \mathbb{R}^{{C/4}\times H \times W}$ that aggregates
multiscale contextual information. In addition, Generalized Mean Pooling
$GeM(\cdot)$ is introduced to perform global aggregation, producing a more robust
scene level representation $\tilde{f}_{i++}^{gd} \in \mathbb{R}^{{C/4}\times 1 \times 1}$,
which is defined as
\begin{equation}
\tilde{f}_{i++}^{d}
=
\theta_{1\times1}(
\mathrm{cat}(
\tilde{f}_{i+}^{d1},
\tilde{f}_{i+}^{d2},
\tilde{f}_{i+}^{d3},
\tilde{f}_{i+}^{d4}
)
),
\end{equation}
\begin{equation}
\tilde{f}_{i++}^{gd}
=
GeM(
\tilde{f}_{i++}^{d}
),
\end{equation}
where GeM enables a continuous transition between average pooling and max pooling
through a learnable exponent $p$, which emphasizes salient response regions during
aggregation. The global feature $\tilde{f}_{i++}^{gd}$ is then broadcast and added
to the feature map to recalibrate local responses. Subsequently, a nonlinear
activation $F_{\mathrm{ReLU}}$ and a $1\times1$ convolution $\theta_{1\times1}$
are applied to expand the channel dimension from $C/4$ back to $C$, yielding the
enhanced feature $\tilde{f}_{i+++}^{d} \in \mathbb{R}^{{C}\times H \times W}$. This
process can be described as
\begin{equation}
\tilde{f}_{i+++}^{d}
=
\theta_{1\times1}(
F_{\mathrm{ReLU}}(
\tilde{f}_{i++}^{d}
+
\tilde{f}_{i++}^{gd}
)
).
\end{equation}

Finally, the enhanced feature $\tilde{f}_{i+++}^{d} \in \mathbb{R}^{C\times H \times W}$
is combined with the original input feature $f_i^{d} \in \mathbb{R}^{C\times H \times W}$
through residual fusion to retain low level details and facilitate stable gradient
propagation, producing the output feature $f_i^{dl} \in \mathbb{R}^{C\times H \times W}$:
\begin{equation}
f_i^{dl}
=
\frac{1}{2}
(
\tilde{f}_{i+++}^{d}
+
f_i^{d}
).
\end{equation}

 \begin{figure*}[t]
  \centering
  \includegraphics[width=7in]{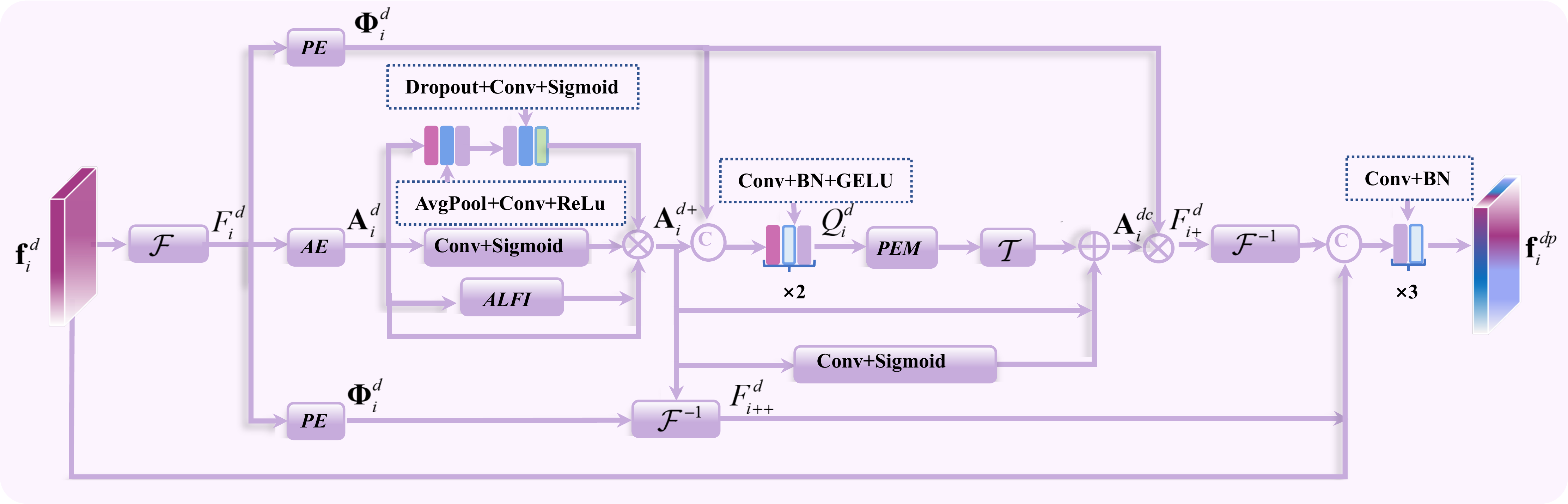}
\caption{Overview of the FSAB. The branch transforms spatial features into the frequency domain and decomposes them into amplitude and phase components. Adaptive frequency reweighting and modulation are applied to the amplitude spectrum, while phase structures are preserved to maintain spatial coherence. The enhanced spectral representations are then projected back to the spatial domain through the inverse Fourier transform, producing frequency-complementary features.}
  \label{fig3}  
  \end{figure*}

\subsection{Frequency Stability Alignment Branch}

As illustrated in Fig.~\ref{fig3}, the core idea of this module is rooted in the complementary nature of frequency and spatial representations. In CVGL, pronounced perspective distortion and non-uniform scale variation make it difficult for a single representation domain to remain reliable across viewpoints. Frequency features emphasize periodic geometric patterns and exhibit relative stability to scale changes through spectral characteristics, whereas spatial features retain precise local texture and structural details. When combined, these two forms of information provide synergistic cues that support more reliable CVGL representation.

Based on this observation, we introduce the FSAB that operates on backbone-extracted feature maps with preserved spatial structure, namely $f_i^{d} \in \mathbb{R}^{C\times H \times W}$ and $f_j^{s} \in \mathbb{R}^{C\times H \times W}$. This branch focuses on extracting frequency domain statistical cues that remain stable across viewpoints and uses them to supplement spatial domain information. Through joint utilization of frequency domain and spatial features, the branch supplies additional constraints for subsequent cross-domain matching and enhances robustness under challenging viewpoint variations.

Similarly, taking the UAV feature $f_i^{d} \in \mathbb{R}^{C\times H \times W}$ as an example, a two-dimensional real valued fast Fourier transform $\mathcal{F}(\cdot)$ is first applied to project the spatial domain feature into the frequency domain, yielding a complex valued spectral representation $F_i^{d} \in \mathbb{C}^{C\times H \times W'}$. Here, $W'=(W/2)+1$ corresponds to the reduced frequency width resulting from the conjugate symmetry property of the real valued FFT. Based on this representation, an amplitude extraction module $AE(\cdot)$ is employed to obtain the amplitude spectrum $A_i^{d}$, while a phase extraction module $PE(\cdot)$ is used to derive the phase spectrum $\Phi_i^{d}$,
\begin{equation}
F_i^{d} = \mathcal{F}(f_i^{d}), \quad
A_i^{d} = AE(F_i^{d}), \quad
\Phi_i^{d} = PE(F_i^{d}),
\end{equation}
where the amplitude spectrum $A_i^{d} \in \mathbb{R}^{C\times H \times W'}$ characterizes the global energy distribution and frequency strength of the image, whereas the phase spectrum $\Phi_i^{d} \in \mathbb{R}^{C\times H \times W'}$ encodes spatial geometric relationships and structural information~\cite{gatys2016image}. In CVGL matching, different frequency components contribute unevenly across both channel and spatial dimensions. We therefore introduce a joint channel--spatial frequency importance mechanism to strengthen the discriminative ability of frequency features. For channel-level modulation, adaptive global average pooling $\Lambda_{2}(\cdot)$ is first applied to aggregate the spatial dimensions of $A_i^{d}$ into a $C\times1\times1$ representation that captures global spectral statistics. A bottleneck mapping composed of two $1\times1$ convolutions $\theta_{1\times1}$ is then used to compress and restore the channel dimension, with a nonlinear activation $F_{\mathrm{ReLU}}$ inserted between them. A Sigmoid function $\delta(\cdot)$ subsequently generates the channel-wise modulation weight $W_c$, formulated as
\begin{equation}
W_c
=
\delta\!\left(
\theta_{1\times1}\!\left(
F_{\mathrm{ReLU}}\!\left(
\theta_{1\times1}\!\left(
\Lambda_{2}\!\left(A_i^{d}\right)
\right)
\right)
\right)
\right).
\end{equation}

Complementary to channel-level modulation, spatial-level modulation focuses on capturing the importance distribution of spectral responses across spatial locations. Specifically, a $3\times3$ convolution $\theta_{3\times3}$ is applied to the amplitude spectrum $A_i^{d}$ to account for spatial neighborhood relationships, followed by a Sigmoid activation $\delta(\cdot)$ to generate the spatial modulation weight $W_s$. In addition, $A_i^{d}$ is processed by an Adaptive Learnable Frequency Importance (ALFI) module $ALFI(\cdot)$, which introduces a learnable parameter $\tau \in \mathbb{R}^{C\times1\times1}$ for further calibration of channel responses. After channel-level modulation, spatial-level modulation, and channel calibration, a weighted amplitude spectrum $A_i^{d+}$ is obtained. This procedure can be expressed as
\begin{equation}
W_s
=
\delta\!\left(
\theta_{3\times3}\!\left(
A_i^{d}
\right)
\right),
\end{equation}
\begin{equation}
\tau
=
ALFI\!\left(
A_i^{d}
\right),
\end{equation}
\begin{equation}
A_i^{d+}
=
\tau\, W_s\, W_c\, A_i^{d}.
\end{equation}

To better utilize stability differences and complementary characteristics under varying viewing conditions, amplitude and phase information are jointly considered in the frequency domain. This combination strengthens the representation of spatial positional relationships and global dependencies within the spectral domain. Specifically, the weighted amplitude $A_i^{d+}$ and the normalized phase $\Phi_i^{d}/\pi$ are concatenated along the channel dimension, after which a $1\times1$ projection convolution $\theta_{1\times1}$ reduces the channel dimension from $2C$ to $C$, followed by normalization $\mathcal{N}(\cdot)$ and a GELU activation $F_{\mathrm{GELU}}$. A $3\times3$ depthwise separable convolution $\theta_{3\times3}$ is then applied to aggregate spatial neighborhood information from both amplitude and phase, again followed by normalization and GELU activation, producing an initial encoded feature $Q_i^{d}$, which is formulated as
\begin{equation}
Q_i^{d}
=
F_{\mathrm{GELU}}\!\left(
\mathcal{N}\!\left(
\theta_{3\times3}\!\left(
F_{\mathrm{GELU}}\!\left(
\mathcal{N}\!\left(
\theta_{1\times1}\!\left(
\mathrm{cat}\!\left(
A_i^{d+},\, \Phi_i^{d}/\pi
\right)
\right)
\right)
\right)
\right)
\right)
\right)
\end{equation}
where $Q_i^{d} \in \mathbb{R}^{C\times H \times W'}$. To incorporate explicit spatial positional information in the frequency domain, continuous normalized coordinates are constructed along the height and width dimensions, denoted as $\mu_i^{d} \in \mathbb{R}^{2\times H \times W'}$. Compared with discrete absolute positional encodings, continuous normalized coordinates provide resolution-invariant relative position cues, which are beneficial for handling inputs at different spatial scales. A positional encoding module $PEM(\cdot)$ then maps $\mu_i^{d}$ into high-dimensional positional embeddings, supplying explicit spatial cues for the subsequent self-attention operation~\cite{liu2021swin}. The encoded feature $Q_i^{d}$ is combined with the positional encoding through residual addition and passed to a multi-head self-attention module $\mathcal{T}(\cdot)$ to capture long range  spectral dependencies, yielding the attention-enhanced feature $F_i^{d+}$,
\begin{equation}
F_i^{d+}
=
\mathcal{T}\!\left(
Q_i^{d}
+
PEM\!\left(
\mu_i^{d}
\right)
\right),
\end{equation}
where $F_i^{d+} \in \mathbb{R}^{C\times H \times W'}$. Although $F_i^{d+}$ incorporates global spectral context through self-attention, the original spectral details may not be fully retained during subsequent spatial domain reconstruction, as self-attention can smooth localized high frequency responses. To alleviate this issue, a residual gating and multipath reconstruction in the frequency domain is introduced to combine the attention-enhanced representation with the original spectral information, allowing global context to be utilized while maintaining local spectral details.

An adaptive residual weight $W_e$ is generated through a $1\times1$ convolution $\theta_{1\times1}$ followed by a Sigmoid activation $\delta(\cdot)$. This weight is derived from the original weighted amplitude $A_i^{d+}$ and allows the fusion ratio to be adjusted according to the characteristics of different frequency components,
\begin{equation}
W_e
=
\delta\!\left(
\theta_{1\times1}\!\left(
A_i^{d+}
\right)
\right),
\end{equation}
where $W_e \in \mathbb{R}^{C\times H \times W'}$. To suppress noise while retaining discriminative frequency information, the attention-enhanced feature $F_i^{d+}$, which is first mapped to a bounded non negative  response via a Sigmoid activation $\delta(\cdot)$, is fused with the original weighted amplitude $A_i^{d+}$ under the guidance of the residual weight $W_e$, yielding the fused amplitude $A_i^{dc}$.

\begin{equation}
A_i^{dc}
=
\delta\!\left(
F_i^{d+}
\right)
\times
(1 - W_e)
+
A_i^{d+}
\times
W_e,
\end{equation}
where $A_i^{dc} \in \mathbb{R}^{C\times H \times W'}$. This weighted fusion mechanism provides a flexible balance between attention-enhanced features and original spectral information~\cite{shi2022accurate}: when $W_e$ approaches $1$, the original amplitude information is largely preserved, whereas when $W_e$ approaches $0$, the fusion emphasizes the attention-enhanced amplitude.

To further integrate complementary cues from the spatial and frequency domains, three parallel reconstruction paths are introduced. By retaining feature representations at different hierarchical levels, the network is able to adaptively select effective combinations of information during reconstruction, which supports robust CVGL matching.

First, the spatial feature $f_i^{d} \in \mathbb{R}^{C\times H \times W}$ produced by the backbone network is directly preserved. This path retains the original spatial domain information without any frequency domain processing, maintaining complete local texture details and serving as a reference feature for subsequent fusion.

Second, the fused amplitude $A_i^{dc}$ is combined with the original phase $\Phi_i^{d}$ to reconstruct a complex valued spectrum, which is then transformed back into the spatial domain through the inverse Fourier transform $\mathcal{F}^{-1}(\cdot)$, yielding a frequency enhanced spatial representation,
\begin{equation}
F_{i+}^{d}
=
\mathcal{F}^{-1}(
A_i^{dc} \cdot e^{j\Phi_i^{d}}
),
\end{equation}
where $F_{i+}^{d} \in \mathbb{R}^{C\times H \times W}$. This reconstruction path exploits the long range dependencies introduced by the self-attention mechanism, allowing global spectral patterns to be reflected in the spatial domain and providing clear advantages in handling scale variation and geometric distortion. Following the same procedure, an alternative reconstruction is obtained by combining the triply gated weighted amplitude $A_i^{d+}$ without attention enhancement with the original phase $\Phi_i^{d}$,
\begin{equation}
F_{i++}^{d}
=
\mathcal{F}^{-1}(
A_i^{d+} \cdot e^{j\Phi_i^{d}}
),
\end{equation}
where $F_{i++}^{d} \in \mathbb{R}^{C\times H \times W}$. This path preserves spectral information without attention modulation, mitigating potential information loss caused by excessive smoothing and providing complementary frequency domain representations.

Finally, the three feature streams are concatenated along the channel dimension and integrated by a fusion module $\mathcal{G}(\cdot)$ to generate the final frequency domain complementary feature $f_i^{dp}$,
\begin{equation}
f_i^{dp}
=
\mathcal{G}\!\left(
\mathrm{cat}\!\left(
f_i^{d},\,
F_{i+}^{d},\,
F_{i++}^{d}
\right)
\right).
\end{equation}

The fusion module $\mathcal{G}$ is composed of three successive $1\times1$ convolutional layers that progressively reduce the channel dimension from $3C$ to $C$. Each layer is followed by batch normalization, GELU activation, and Dropout regularization. Through this progressive integration, the fusion module learns to balance the three feature streams and selectively retain information that is most effective for CVGL matching. As a result, $f_i^{dp} \in \mathbb{R}^{C\times H \times W}$, and $f_j^{sp}$ is obtained in the same manner.

\subsection{Loss Optimization}
\label{subsec1}
The training framework employs multiple loss functions to jointly guide network optimization from complementary perspectives. Specifically, the global semantic features $f_i^{dg}$ and $f_j^{sg}$ produced by the GSCB are supervised by the cross-entropy loss $L_{CCE}$, promoting class separability and enhancing global semantic discrimination. In the LGSB, the InfoNCE loss $L_{InfoNCE}$ is applied to the multiscale features $f_i^{dl}$ and $f_j^{sl}$, encouraging positive cross-view pairs to remain close while separating negative samples in the embedding space and supporting stable semantic correspondence across viewpoints. Additionally, the frequency-enhanced features $f_i^{dp}$ and $f_j^{sp}$ generated by the FSAB are optimized using the domain and spatial alignment loss $L_{DSA}$, which performs contrastive supervision on the reconstructed spatial representations and encourages consistency under viewpoint changes and geometric perturbations. The overall objective minimizes a weighted combination of these loss terms to balance discrimination and cross-domain robustness, which is expressed as
\begin{equation}
\mathcal{L}_{\text{total}}
=
\lambda_1 \mathcal{L}_{\text{CE}}
+
\lambda_2 \mathcal{L}_{\text{InfoNCE}}
+
\lambda_3 \mathcal{L}_{\text{DSA}},
\end{equation}
where $\lambda_1 = 0.1$, $\lambda_2 = 1.0$, and $\lambda_3 = 1.3$ control the relative contribution of each loss component. Notably, the frequency domain alignment loss term is assigned a higher weight than the global classification loss, which reflects the importance of frequency domain stability when handling cross-view geometric asymmetry. Through this multilevel supervision, the network is guided to learn complementary properties related to global semantics, local geometric discrimination, and statistical consistency in the frequency domain, supporting robust CVGL matching under large viewpoint variations and challenging imaging conditions.

\begin{table}[t]
  \centering
  \caption{Comparisons between the proposed method and some state-of-the-art methods on the University-1652 datasets. The best results are highlighted in red, while the second-best results are highlighted in blue.}
  \label{tab:comparison}
  \small
  \resizebox{\columnwidth}{!}{ 
  \begin{tabular}{cccccc}
  \hline
   &  & \multicolumn{2}{c}{Drone$\rightarrow$Satellite}  & \multicolumn{2}{c}{Satellite$\rightarrow$Drone}  \\ \cline{3-6}
  \multirow{-2}{*}{Model} & \multirow{-2}{*}{Venue} & {R@1} & {AP} & {R@1} & {AP} \\ \hline
  MuSe-Net\cite{wang2024multiple}      & PR’2024      & 74.48       & 77.83      & 88.02     & 75.10 \\
  LPN\cite{wang2021each}               & TCSVT’2021   & 75.93       & 79.14      & 86.45     & 74.49 \\
  F3-Net\cite{sun2023f3}               & TGRS’2023    & 78.64       & 81.60      & -         & -     \\
  TransFG\cite{zhao2024transfg}        & TGRS’2024    & 84.01       & 86.31      & 90.16     & 84.61 \\
  IFSs\cite{ge2024multibranch}         & TGRS’2024    & 86.06       & 88.08      & 91.44     & 85.73 \\
  MCCG\cite{shen2023mccg}              & TCSVT’2023   & 89.40       & 91.07      & 95.01     & 89.93 \\
  SDPL\cite{chen2024sdpl}              & TCSVT’2024   & 90.16       & 91.64      & 93.58     & 89.45 \\
  MFJR\cite{ge2024multi}               & TGRS’2024    & 91.87       & 93.15      & 95.29     & 91.51 \\
  CCR\cite{du2024ccr}                  & TCSVT2024    & 92.54       & 93.78      & 95.15     & 91.80 \\
 ViT-SegMatchNet\cite{ZENG2025804}& ISPRS'2025     & 92.60       & 93.80      & 95.59     & 92.30 \\
  Sample4Geo\cite{deuser2023sample4geo}& ICCV’2023    & 92.65       & 93.81      & 95.14     & 91.39 \\
  SRLN\cite{lv2024direction}           & TGRS’2024    & 92.70       & 93.77      & 95.14     & 91.97 \\
   MEAN\cite{chen2025multilevel}                                  & TGRS’2025    & 93.55       & 94.53      & 96.01     & 92.08 \\
  DAC\cite{xia2024enhancing}           & TCSVT2024    & \textcolor{red}{94.67} & \textcolor{red}{95.50} & \textcolor{blue}{96.43} & \textcolor{red}{93.79} \\
  \hline
  SFDE (Ours)                            &              & \textcolor{blue}{93.75} & \textcolor{blue}{94.72} & \textcolor{red}{96.72} & \textcolor{blue}{92.40} \\
  \hline
  \end{tabular}}
\end{table}

\section{Experimental Results}
\subsection{Experimental Datasets and Evaluation Metrics}

We conduct systematic evaluations on three representative CVGL benchmark datasets. These datasets span diverse viewpoint discrepancies, spatial scales, and imaging conditions, enabling a comprehensive assessment of the proposed SFDE framework.

University-1652~\cite{zheng2020university} serves as a core benchmark for the CVGL community. It contains 1,652 geo-locations from 72 universities with UAV, satellite, and ground views. The training set includes 701 locations from 33 universities, while the test set consists of 951 locations from 39 universities with non-overlapping geographic splits. This dataset introduces UAV viewpoints into CVGL research and allows evaluation under multi-view settings.

SUES-200~\cite{zhu2023sues} emphasizes altitude variation. It contains 200 geo-locations, with 120 used for training and 80 reserved for testing. Each location includes one satellite image and a sequence of UAV images captured at four altitudes (150 m, 200 m, 250 m, and 300 m). This dataset is commonly used to examine robustness under large scale changes across different environments, such as parks, lakes, and building clusters.

Multi-weather University-1652~\cite{wang2024multiple} extends University-1652 by introducing ten simulated weather conditions based on physics driven rendering. It provides a standardized evaluation setting under adverse imaging conditions, including illumination variation and texture degradation, particularly relevant for assessing statistical stability in the frequency domain.

Performance is evaluated using standard metrics widely adopted in the CVGL community~\cite{hu2022beyond}. Recall@K (R@K) measures the proportion of queries for which the correct match appears within the top-$K$ retrieval results. Average Precision (AP) summarizes retrieval quality by jointly considering precision and recall across different ranking thresholds.

\begin{table*}[t]
\centering
\caption{Comparison with state-of-the-art results under multi-weather conditions on the University-1652 dataset. The best results are highlighted in red, while the second-best results are highlighted in blue.}
\label{tab:Multi-weather University-1652}
\resizebox{\textwidth}{!}{
\begin{tabular}{lcccccccccc}
\hline
\multirow{2}{*}{Model} & Normal & Fog & Rain & Snow & Fog+Rain & Fog+Snow & Rain+Snow & Dark & Over-exposure& Wind \\ \cline{2-11}
 & R@1/AP & R@1/AP & R@1/AP & R@1/AP & R@1/AP & R@1/AP & R@1/AP & R@1/AP & R@1/AP & R@1/AP \\ \hline
\multicolumn{11}{c}{Drone$\rightarrow$Satellite} \\ \hline
LPN\cite{wang2021each} & 74.33/77.60 & 69.31/72.95 & 67.96/71.72 & 64.90/68.85 & 64.51/68.52 & 54.16/58.73 & 65.38/69.29 & 53.68/58.10 & 60.90/65.27 & 66.46/70.35 \\
MuSeNet\cite{wang2024multiple} & 74.48/77.83 & 69.47/73.24 & 70.55/74.14 & 65.72/69.70 & 65.59/69.64 & 54.69/59.24 & 65.64/70.54 & 53.85/58.49 & 61.65/65.51 & 69.45/73.22 \\
Sample4Geo\cite{deuser2023sample4geo} & 90.55/92.18 & 89.72/91.48 & 85.89/88.11 & 86.64/88.18 & 85.88/88.16 & 84.64/87.11 & 85.98/88.16 & 87.90/89.87 & 76.72/80.18 & 83.39/89.51 \\
MEAN\cite{chen2025multilevel} & {\color{blue}90.81}/{\color{blue}92.32} & {\color{blue}90.97}/{\color{blue}92.52} & {\color{blue}88.19}/{\color{blue}90.05} & {\color{blue}88.69}/{\color{blue}90.49} & {\color{blue}86.75}/{\color{blue}88.84} & {\color{blue}86.00}/{\color{blue}88.22} & {\color{blue}87.21}/{\color{blue}89.21} & {\color{blue}87.90}/{\color{blue}89.87} & {\color{red}80.54}/{\color{red}83.53} & {\color{blue}89.27}/{\color{blue}91.01} \\
SFDE (Ours) & {\color{red}92.99}/{\color{red}94.22} & {\color{red}93.33}/{\color{red}94.50} & {\color{red}92.99}/{\color{red}94.20} & {\color{red}93.20}/{\color{red}94.34} & {\color{red}93.35}/{\color{red}94.50} & {\color{red}92.78}/{\color{red}93.99} & {\color{red}92.76}/{\color{red}94.00} & {\color{red}90.30}/{\color{red}91.89} & {\color{blue}78.59}/{\color{blue}81.66} & {\color{red}90.18}/{\color{red}91.81} \\ \hline
\multicolumn{11}{c}{Satellite$\rightarrow$Drone} \\ \hline
LPN\cite{wang2021each} & 87.02/75.19 & 86.16/71.34 & 83.88/69.49 & 82.88/65.39 & 84.59/66.28 & 79.60/55.19 & 84.17/66.26 & 82.88/52.05 & 81.03/62.24 & 84.14/67.35 \\
MuSeNet\cite{wang2024multiple} & 88.02/75.10 & 87.87/69.85 & 87.73/71.12 & 83.74/66.52 & 85.02/67.78 & 80.88/54.26 & 84.88/67.75 & 80.74/53.01 & 81.60/62.09 & 86.31/70.03 \\
Sample4Geo\cite{deuser2023sample4geo} & 95.86/89.86 & 95.72/88.95 & 94.44/85.71 & 95.01/86.73 & 93.44/85.27 & 93.72/84.78 & 93.15/85.50 & 96.01/87.06 & 89.87/74.52 & 95.29/87.06 \\
MEAN\cite{chen2025multilevel} & {\color{blue}96.58}/{\color{blue}89.93} & {\color{blue}96.00}/{\color{blue}89.49} & {\color{blue}95.15}/{\color{blue}88.87} & {\color{blue}94.44}/{\color{blue}87.44} & {\color{blue}93.58}/{\color{blue}86.91} & {\color{blue}94.44}/{\color{blue}87.44} & {\color{blue}93.72}/{\color{blue}86.91} & {\color{blue}96.29}/{\color{blue}89.87} & {\color{blue}92.87}/{\color{red}79.66} & {\color{blue}95.44}/{\color{blue}87.06} \\
SFDE (Ours) & {\color{red}97.15}/{\color{red}92.27} & {\color{red}97.00}/{\color{red}92.03} & {\color{red}97.00}/{\color{red}92.42} & {\color{red}96.72}/{\color{red}92.48} & {\color{red}97.00}/{\color{red}92.50} & {\color{red}96.86}/{\color{red}91.46} & {\color{red}97.15}/{\color{red}92.64} & {\color{red}96.43}/{\color{red}90.74} & {\color{red}93.01}/{\color{blue}77.48} & {\color{red}96.58}/{\color{red}89.81} \\ \hline
\end{tabular}}
\end{table*}
\begin{table*}[t]
  \centering
  \tiny
  \caption{Comparisons between the proposed method and some state-of-the-art methods on the SUES-200 dataset (Drone$\rightarrow$Satellite). The best results are in red, the second-best are in blue.}
  \label{tab:comparison_SUES200_1}
  \resizebox{\textwidth}{!}{
  \begin{tabular}{lcccccccccc}
  \hline
  \multirow{3}{*}{Model} & \multirow{3}{*}{Venue} & \multicolumn{8}{c}{Drone$\rightarrow$Satellite} \\ \cline{3-10}
   &  & \multicolumn{2}{c}{150m} & \multicolumn{2}{c}{200m} & \multicolumn{2}{c}{250m} & \multicolumn{2}{c}{300m} \\ \cline{3-10}
   &  & R@1 & AP & R@1 & AP & R@1 & AP & R@1 & AP \\ \hline
  LPN\cite{wang2021each}                    & TCSVT'2022  & 61.58 & 67.23 & 70.85 & 75.96 & 80.38 & 83.80 & 81.47 & 84.53 \\
  IFSs\cite{ge2024multibranch}              & TGRS'2024   & 77.57 & 81.30 & 89.50 & 91.40 & 92.58 & 94.21 & 97.40 & 97.92 \\
  MCCG\cite{shen2023mccg}                   & TCSVT'2023  & 82.22 & 85.47 & 89.38 & 91.41 & 93.82 & 95.04 & 95.07 & 96.20 \\
  SDPL\cite{chen2024sdpl}                   & TCSVT'2024  & 82.95 & 85.82 & 92.73 & 94.07 & 96.05 & 96.69 & 97.83 & 98.05 \\
  CCR\cite{du2024ccr}                       & TCSVT'2024  & 87.08 & 89.55 & 93.57 & 94.90 & 95.42 & 96.28 & 96.82 & 97.39 \\
  MFJR\cite{ge2024multi}                    & TGRS'2024   & 88.95 & 91.05 & 93.60 & 94.72 & 95.42 & 96.28 & 97.45 & 97.84 \\
  SRLN\cite{lv2024direction}                & TGRS'2024   & 89.90 & 91.90 & 94.32 & 95.65 & 95.92 & 96.79 & 96.37 & 97.21 \\
  Sample4Geo\cite{deuser2023sample4geo}     & ICCV'2023   & 92.60 & 94.00 & 97.38 & 97.81 & 98.28 & 98.64 & 99.18 & 99.36 \\
   MEAN\cite{chen2025multilevel}                                       & TGRS'2025           & 95.50 & 96.46 & \textcolor{blue}{98.38} & \textcolor{blue}{98.72} & \textcolor{blue}{98.95} & \textcolor{blue}{99.17}  & 99.52  & 99.63 \\
  ViT-SegMatchNet\cite{ZENG2025804}& ISPRS'2025   & 95.78 & 96.62 & 97.86 & 98.30 & 98.90 & 99.14 & \textcolor{blue}{99.76} & \textcolor{blue}{99.82} \\
  DAC\cite{xia2024enhancing}                & TCSVT'2024  & \textcolor{red}{96.80}  & \textcolor{red}{97.54} & 97.48 & 97.97 & 98.20 & 98.62 & 97.58 & 98.14 \\
  \hline
  SFDE (Ours)                                 & -           & \textcolor{blue}{95.90} & \textcolor{blue}{96.67} & \textcolor{red}{98.55} & \textcolor{red}{98.87} & \textcolor{red}{99.73} & \textcolor{red}{99.79}  & \textcolor{red}{99.98}  & \textcolor{red}{99.98} \\\hline
  \end{tabular}}
\end{table*}

\subsection{Implementation Details}

We construct training batches using a symmetric sampling strategy, where each batch contains 32 UAV images and 32 satellite images. The ConvNeXt-Tiny backbone is initialized with ImageNet-pretrained weights, while the newly added classifiers are initialized using Kaiming initialization. All images are uniformly resized to a resolution of $384 \times 384$ and augmented using random cropping, random horizontal flipping, and random rotation. Optimization is performed using the AdamW optimizer with an initial learning rate of 0.001. A cosine annealing learning rate scheduler is employed, with the warm-up phase accounting for 10\% of the total training steps. The hyperparameters of the SFDE loss function are set to $\lambda_1 = 0.1$, $\lambda_2 = 1.0$, and $\lambda_3 = 1.3$. All experiments are implemented using the PyTorch framework and conducted on an Ubuntu 22.04 system equipped with an NVIDIA RTX 4090 GPU.

\begin{table*}[ht]
  \centering
  \tiny
  \caption{{\color{black}Comparisons between the proposed method and some state-of-the-art methods on the SUES-200 dataset in the Satellite$\rightarrow$Drone. The best results are highlighted in red, while the second-best results are highlighted in blue.}}
  \label{tab:comparison SUES-200-2}
  \resizebox{\textwidth}{!}{
  \begin{tabular}{cccccccccc}
  \hline
   &  & \multicolumn{8}{c}{Satellite$\rightarrow$Drone} \\ \cline{3-10} 
   &  & \multicolumn{2}{c}{150m} & \multicolumn{2}{c}{200m} & \multicolumn{2}{c}{250m} & \multicolumn{2}{c}{300m} \\ \cline{3-10} 
  \multirow{-3}{*}{Model} & \multirow{-3}{*}{Venue} & R@1 & AP & R@1 & AP & R@1 & AP & R@1 & AP \\ \hline
  LPN\cite{wang2021each}       & TCSVT'2022       & 83.75 & 83.75 & 83.75 & 83.75 & 83.75 & 83.75 & 83.75 & 83.75 \\
  CCR\cite{du2024ccr}          & TCSVT'2024       & 92.50 & 88.54 & 97.50 & 95.22 & 97.50 & 97.10 & 97.50 & 97.49 \\
  IFSs\cite{ge2024multibranch} & TGRS'2024        & 93.75 & 79.49 & 97.50 & 90.52 & 97.50 & 96.03 & \textcolor{red}{100.00} & 97.66 \\
  MCCG\cite{shen2023mccg}      & TCSVT'2023       & 93.75 & 89.72 & 93.75 & 92.21 & 96.25 & 96.14 & 98.75 & 96.64 \\
  SDPL\cite{chen2024sdpl}      & TCSVT'2024       & 93.75 & 83.75 & 96.25 & 92.42 & 97.50 & 95.65 & 96.25 & 96.17 \\
  SRLN\cite{lv2024direction}   & TGRS'2024        & 93.75 & 93.01 & 97.50 & 95.08 & 97.50 & 96.52 & 97.50 & 96.71 \\
  MFJR\cite{ge2024multi}       & TGRS'2024        & 95.00 & 89.31 & 96.25 & 94.72 & 94.69 & 96.92 & 98.75 & 97.14 \\
Sample4Geo\cite{deuser2023sample4geo} & ICCV'2023        &97.50 &93.63 &\textcolor{blue}{98.75} &96.70 &\textcolor{blue}{98.75} &98.28 &98.75 &98.05\\
  DAC\cite{xia2024enhancing}   & TCSVT'2024       &97.50 &94.06  & \textcolor{blue}{98.75} & 96.66 &\textcolor{blue}{98.75} & 98.09 &98.75 & 97.87 \\
  MEAN\cite{chen2025multilevel}& TGRS’2025                &97.50  & \textcolor{blue}{94.75} &\textcolor{red}{100.00}  & 97.09 & \textcolor{red}{100.00} &98.28  & \textcolor{red}{100.00} &\textcolor{blue}{99.21} \\
  ViT-SegMatchNet\cite{ZENG2025804}& ISPRS'2025                &\textcolor{blue}{97.88}  & \textcolor{red}{95.28} &98.50  & \textcolor{blue}{97.41} & \textcolor{blue}{98.75} &\textcolor{blue}{98.44}  & \textcolor{blue}{98.88} &98.62 \\
\hline
  SFDE(Ours) & -                &\textcolor{red}{98.75}  & 94.71 &\textcolor{red}{100.00}  & \textcolor{red}{98.53} & \textcolor{red}{100.00} &\textcolor{red}{99.56}  & \textcolor{red}{100.00} &\textcolor{red}{99.65} \\ \hline
  \end{tabular}}
\end{table*}

\begin{table*}[t]
\centering
\tiny
\caption{Comparisons between the proposed method and state-of-the-art methods in cross-domain evaluation on Drone$\rightarrow$Satellite. The best results are highlighted in red, while the second-best results are highlighted in blue.}
\label{tab:comparison University-SUES-1}
\resizebox{\textwidth}{!}{
\begin{tabular}{cccccccccc}
\hline
 &  & \multicolumn{8}{c}{Drone$\rightarrow$Satellite} \\ \cline{3-10} 
 &  & \multicolumn{2}{c}{150m} & \multicolumn{2}{c}{200m} & \multicolumn{2}{c}{250m} & \multicolumn{2}{c}{300m} \\ \cline{3-10} 
\multirow{-3}{*}{Model} & \multirow{-3}{*}{Venue} & R@1 & AP & R@1 & AP & R@1 & AP & R@1 & AP \\ \hline
MCCG\cite{shen2023mccg}      & TCSVT'2023 & 57.62 & 62.80 & 66.83 & 71.60 & 74.25 & 78.35 & 82.55 & 85.27  \\
Sample4Geo\cite{deuser2023sample4geo}& ICCV'2023  & 70.05 & 74.93 & 80.68 & 83.90 & 87.35 & 89.72 & 90.03 & 91.91   \\
DAC\cite{xia2024enhancing}       & TCSVT'2024 & 76.65 & 80.56 & 86.45 & 89.00 &\textcolor{blue}{92.95}  & \textcolor{blue}{94.18} & \textcolor{blue}{94.63} & 95.45   \\
 MEAN\cite{chen2025multilevel}& TGRS’2025          & \textcolor{blue}{81.73} & \textcolor{red}{85.72}  & \textcolor{red}{89.05}  & \textcolor{red}{91.00}  & 92.13  & 93.60  & 94.53  &\textcolor{blue}{95.76}    \\\hline
SFDE(Ours)& -          & \textcolor{red}{82.83}&\textcolor{blue}{85.34}&\textcolor{blue}{88.58}&\textcolor{blue}{90.34}&\textcolor{red}{93.23}&\textcolor{red}{94.33}&\textcolor{red}{95.38}&\textcolor{red}{96.09}\\ \hline
\end{tabular}}
\end{table*}

\begin{table*}[ht]
\centering
\tiny
\caption{Comparisons between the proposed method and state-of-the-art methods in cross-domain evaluation on Satellite$\rightarrow$Drone. The best results are highlighted in red, while the second-best results are highlighted in blue.}
\label{tab:comparison University-SUES-2}
\resizebox{\textwidth}{!}{
\begin{tabular}{cccccccccc}
\hline
 &  & \multicolumn{8}{c}{Satellite$\rightarrow$Drone} \\ \cline{3-10} 
 &  & \multicolumn{2}{c}{150m} & \multicolumn{2}{c}{200m} & \multicolumn{2}{c}{250m} & \multicolumn{2}{c}{300m} \\ \cline{3-10} 
\multirow{-3}{*}{Model} & \multirow{-3}{*}{Venue} & R@1 & AP & R@1 & AP & R@1 & AP & R@1 & AP \\ \hline
MCCG\cite{shen2023mccg}    & TCSVT'2023  & 61.25 & 53.51 & 82.50 & 67.06 & 81.25 & 74.99 & 87.50 & 80.20   \\
Sample4Geo\cite{deuser2023sample4geo} & ICCV'2023  & 83.75 & 73.83 & 91.25 & 83.42 & \textcolor{blue}{93.75} & 89.07 & \textcolor{blue}{93.75} & 90.66   \\
DAC\cite{xia2024enhancing}        & TCSVT'2024 & 87.50 & 79.87 & \textcolor{red}{96.25} & 88.98  & \textcolor{red}{96.25} & \textcolor{blue}{92.81} & \textcolor{red}{96.25} & 94.00   \\
MEAN\cite{chen2025multilevel}& TGRS'2025& \textcolor{blue}{91.25}   & \textcolor{blue}{81.50}   & \textcolor{red}{96.25}   & \textcolor{blue}{89.55}  & \textcolor{red}{96.25} & 92.36 & \textcolor{red}{96.25}  & \textcolor{red}{94.32}   \\\hline
SFDE(Ours)& -          & \textcolor{red}{92.50}& \textcolor{red}{84.74}& \textcolor{blue}{93.75}&  \textcolor{red}{90.09}  & \textcolor{red}{96.25} & \textcolor{red}{92.97} & \textcolor{red}{96.25}  & \textcolor{blue}{94.12}   \\ \hline
\end{tabular}}
\end{table*}

\subsection{Comparison with State-of-the-Art Methods}

\begin{figure}[ht]
    \centering
    \includegraphics[width=3.5in]{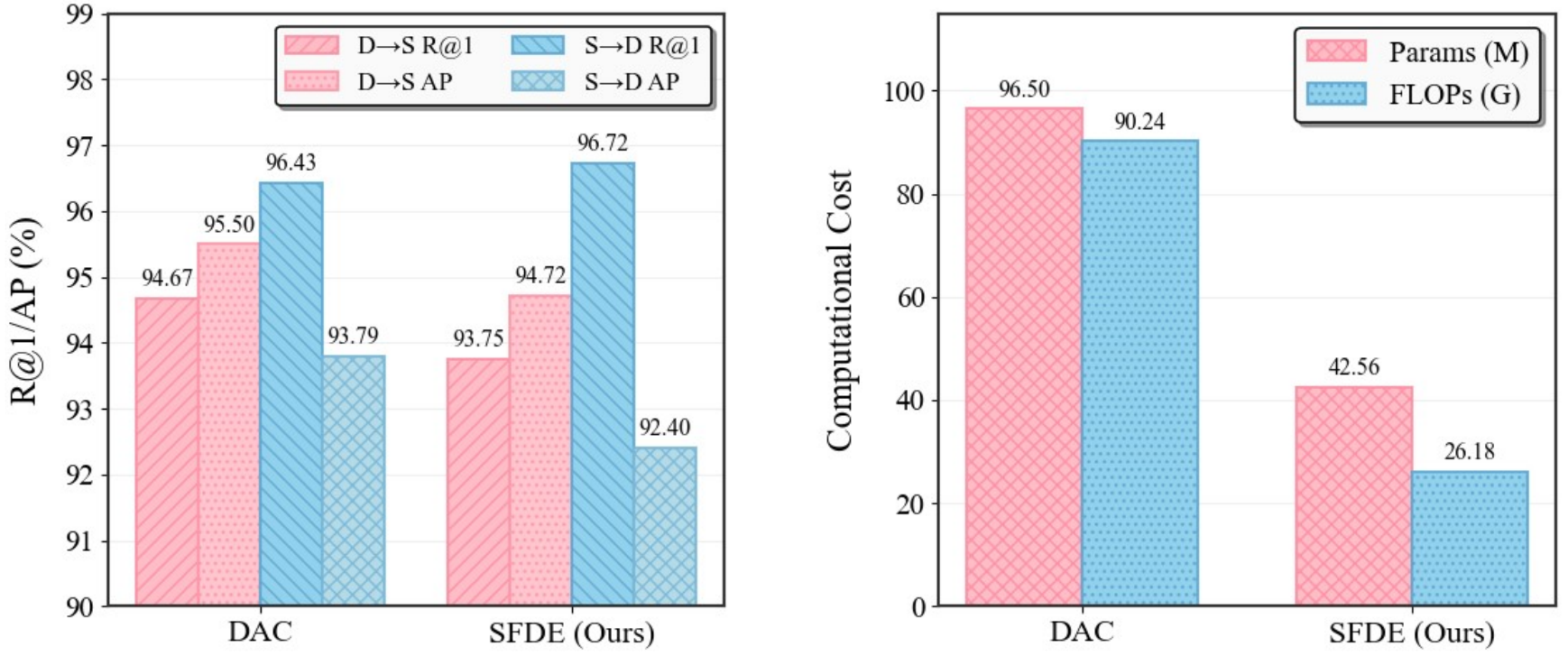}
    \caption{Comparison of computational cost (Params, FLOPs) and performance between DAC and SFDE.}
    \label{fig4}  
\end{figure}

 We compare SFDE against representative state-of-the-art methods in Table~\ref{tab:comparison}. On the Drone$\rightarrow$Satellite task, SFDE achieves 93.75\% R@1 and 94.72\% AP. Despite using a lightweight backbone, it outperforms LPN, MCCG, and SDPL by notable margins. This result indicates that strong performance can be attained without relying on complex architectures. Although DAC achieves slightly better performance on the Drone$\rightarrow$Satellite task, as shown in Fig.~\ref{fig4}, SFDE reduces the parameter count by 55.9\% (42.56M vs.~96.50M) and the computational cost by 71.0\% (26.18G vs.~90.24G FLOPs), leading to a more favorable balance between efficiency and performance. On the Satellite$\rightarrow$Drone task, SFDE surpasses DAC with an R@1 of 96.72\% compared to 96.43\%, highlighting its strong performance under a lightweight architecture and its suitability for deployment in resource constrained edge computing scenarios.

Performance under ten distinct weather conditions is presented in Table~\ref{tab:Multi-weather University-1652}. Under the Drone$\rightarrow$Satellite setting, SFDE achieves the highest R@1 and AP scores in 9 out of the 10 weather conditions. Under the Satellite$\rightarrow$Drone setting, SFDE attains the best R@1 performance across all ten weather conditions. These results demonstrate that the frequency-enhancement branch maintains effectiveness under challenging conditions such as texture degradation, low illumination, and compound environmental interference.

We evaluate performance at four different flight altitudes, as detailed in Tables~\ref{tab:comparison_SUES200_1} and~\ref{tab:comparison SUES-200-2}. On the Drone$\rightarrow$Satellite task, SFDE achieves the highest R@1 and AP at altitudes of 200~m, 250~m, and 300~m. At 150~m altitude, SFDE achieves competitive results with less than 1\% gap to the best-performing method. On the Satellite$\rightarrow$Drone task, SFDE achieves the best R@1 across all four flight altitudes. These results indicate that the integrated design of SFDE maintains stable performance under multiscale conditions and nonlinear viewpoint variations.

\subsection{Comparison with State-of-the-Art Methods on Cross-Domain Generalization Performance}

To further evaluate generalization under distribution shift, we conduct cross-domain experiments by training SFDE on University-1652 and performing zero shot testing on SUES-200, which differs substantially in viewpoint, scale, and environmental characteristics. This setting reflects the common discrepancy between training and deployment domains in real-world UAV applications and provides a meaningful benchmark for examining robustness and cross-view consistency in CVGL models.

In the Drone$\rightarrow$Satellite setting, SFDE achieves the best AP at altitudes of 250~m and 300~m, with particularly pronounced performance gains in the 200~m--300~m range (Tables~\ref{tab:comparison University-SUES-1} and~\ref{tab:comparison University-SUES-2}). In the Satellite$\rightarrow$Drone setting, SFDE consistently outperforms all competing methods in terms of R@1 across all four altitudes, reaching the highest value of 96.25\% at both 250~m and 300~m. In contrast, most methods that rely on complex architectures experience varying degrees of performance degradation under cross-domain conditions.

\subsection{Ablation Studies}

To assess the contribution of individual modules, we conduct ablation studies summarized in Table~\ref{tab:hp_univ1652_1}. The baseline model achieves R@1 scores of 84.00\% and 92.29\% on the Drone$\rightarrow$Satellite and Satellite$\rightarrow$Drone settings, respectively. Incorporating LGSB improves performance to 90.74\% and 95.01\%, yielding R@1 gains of 6.74\% and 2.72\% respectively. This indicates that local contrastive loss enhances sensitivity to fine-grained geometric cues. Adding GSCB to LGSB further improves R@1 to 91.55\% on the Drone$\rightarrow$Satellite task. This suggests that global semantic supervision strengthens topological awareness at broader spatial scales. The LGSB+FSAB configuration attains R@1 scores of 92.21\% and 96.01\%, providing additional improvements of 1.47\% and 1.00\% over LGSB alone, reflecting the complementary role of frequency domain information through statistical stability. The complete SFDE framework achieves the best overall performance, with 93.75\% R@1 and 94.72\% AP on the Drone$\rightarrow$Satellite task, and 96.72\% R@1 and 92.40\% AP on the Satellite$\rightarrow$Drone task. Relative to the baseline, SFDE improves R@1 by 9.75\% and 4.43\% on the two tasks, respectively, indicating clear complementarity among the three branches.

\begin{table}
\centering
\small

\caption{The influence of each component on the performance of proposed method.\\ The best results are highlighted in red.}
\label{tab:hp_univ1652_1}
\resizebox{\columnwidth}{!}{ 
\begin{tabular}{ccccccc} 
\hline
\multicolumn{3}{c}{\multirow{2}{*}{Setting}} & \multicolumn{4}{c}{University-1652}                                                                \\ 
\cline{4-7}
\multicolumn{3}{c}{}                         & \multicolumn{2}{c}{Drone$\rightarrow$Satellite} & \multicolumn{2}{c}{Satellite$\rightarrow$Drone}  \\ 
\hline
LGSB & GSCB & FSAB
                 & R@1                    & AP                     & R@1                    & AP                      \\ \hline
        &             &                         & 84.00                  & 86.51                  & 92.29                  & 82.90                  \\
$\checkmark$  &             &                         & 90.74                  & 92.29                  & 95.01                  & 91.09                  \\
$\checkmark$  & $\checkmark$ &                         & 91.55                  & 93.01                  & 94.72                  & 90.44                  \\
$\checkmark$  &  &$\checkmark$                         & 92.21                  & 93.52                  & 96.01                  & 91.61                  \\
$\checkmark$  & $\checkmark$ & $\checkmark$            & \textcolor{red}{93.75} & \textcolor{red}{94.72} & \textcolor{red}{96.72} & \textcolor{red}{92.40} \\
\hline
\end{tabular}}
\end{table}

\begin{table}
\centering
\tiny
\caption{Effect of different GSCB loss weights on model performance with the LGSB and FSAB loss weights fixed. The best results are highlighted in red, while the second-best results are highlighted in blue.}
\label{tab:hp_univ1652_2}
\resizebox{\columnwidth}{!}{ 
\begin{tabular}{ccccccc}
\hline
\multicolumn{3}{c}{\multirow{2}{*}{Setting}} & \multicolumn{4}{c}{University-1652}                                                                \\ 
\cline{4-7}
\multicolumn{3}{c}{}                         & \multicolumn{2}{c}{Drone$\rightarrow$Satellite} & \multicolumn{2}{c}{Satellite$\rightarrow$Drone}  \\ 
\hline
$\lambda_1$ & $\lambda_2$ & $\lambda_3$              & R@1                    & AP                     & R@1                    & AP                      \\ \hline

0.2    & 1.0        & 1.3                    & \textcolor{blue}{92.23}  & \textcolor{blue}{93.51}  & 94.72 & \textcolor{blue}{90.40}  \\
0.3    & 1.0        & 1.3                    & 91.67                  & 92.99                  & 94.86                  & 90.22                  \\
0.4    & 1.0        & 1.3                    & 91.13                  & 92.49                  & \textcolor{blue}{95.01}                  & 90.15                  \\
0.5    & 1.0        & 1.3                    & 90.64                  & 92.07                  & 94.29                  & 88.71                  \\
0.1    & 1.0        & 1.3                    & \textcolor{red}{93.75} & \textcolor{red}{94.72} & \textcolor{red}{96.72} & \textcolor{red}{92.40}  \\
\hline

\end{tabular}}
\end{table}

\begin{table}
\centering
\caption{Effect of different LGSB loss weights on model performance with the GSCB and FSAB loss weights fixed. The best results are highlighted in red, while the second-best results are highlighted in blue.}
\label{tab:hp_univ1652_3}
\tiny
\resizebox{\columnwidth}{!}{ 
\begin{tabular}{ccccccc} 
\hline
\multicolumn{3}{c}{\multirow{2}{*}{Setting}} & \multicolumn{4}{c}{University-1652}                                                                \\ 
\cline{4-7}
\multicolumn{3}{c}{}                         & \multicolumn{2}{c}{Drone$\rightarrow$Satellite} & \multicolumn{2}{c}{Satellite$\rightarrow$Drone}  \\ 
\hline
$\lambda_1$ & $\lambda_2$ & $\lambda_3$                 & R@1                    & AP                     & R@1                    & AP                      \\ \hline
0.1    & 0.8        & 1.3                    & 92.25                  & 93.49                  & 95.72                  & 91.72                  \\
0.1    & 0.9        & 1.3                    & 92.92                  & 94.04                  & 94.86                  & \textcolor{blue}{91.82}                  \\
0.1    & 1.1        & 1.3                    & \textcolor{blue}{93.10}                   & \textcolor{blue}{94.14}                  & \textcolor{blue}{96.15} &\textcolor{red}{92.40}\\
0.1    & 1.2        & 1.3                    & 92.64                  & 93.78                  & 95.44                  & 91.64                  \\
0.1    & 1.0        & 1.3                    & \textcolor{red}{93.75} & \textcolor{red}{94.72} &  \textcolor{red}{96.72} & \textcolor{red}{92.40}                    \\
\hline
\end{tabular}}
\end{table}

\begin{table}
\centering
\tiny
\caption{Effect of different FSAB loss weights on model performance with the GSCB and LGSB loss weights fixed. The best results are highlighted in red, while the second-best results are highlighted in blue.}
\label{tab:hp_univ1652}
\resizebox{\columnwidth}{!}{ 
\begin{tabular}{ccccccc} 
\hline
\multicolumn{3}{c}{\multirow{2}{*}{Setting}} & \multicolumn{4}{c}{University-1652}                                                                \\ 
\cline{4-7}
\multicolumn{3}{c}{}                         & \multicolumn{2}{c}{Drone$\rightarrow$Satellite} & \multicolumn{2}{c}{Satellite$\rightarrow$Drone}  \\ 
\hline
$\lambda_1$ & $\lambda_2$ & $\lambda_3$                 & R@1                    & AP                     & R@1                    & AP                      \\ \hline
0.1    & 1.0        & 1.1                    & 93.16                  & 94.26                  & \textcolor{blue}{96.43}                  & \textcolor{red}{92.49}                  \\
0.1    & 1.0        & 1.2                    & 92.83                  & 93.94                  & 95.15                  & 92.07                  \\
0.1    & 1.0        & 1.4                    & \textcolor{blue}{93.34}                  & \textcolor{blue}{94.32}                  & 95.72                  & 92.28                  \\
0.1    & 1.0        & 1.5                    & 93.25                  & 94.29                  & 95.72                  & 92.35                  \\
0.1    & 1.0        & 1.3                    & \textcolor{red}{93.75} & \textcolor{red}{94.72} & \textcolor{red}{96.72} & \textcolor{blue}{92.40}  \\
\hline
\end{tabular}}
\end{table}

\begin{figure}[ht]
    \centering
    \includegraphics[width=3.5in]{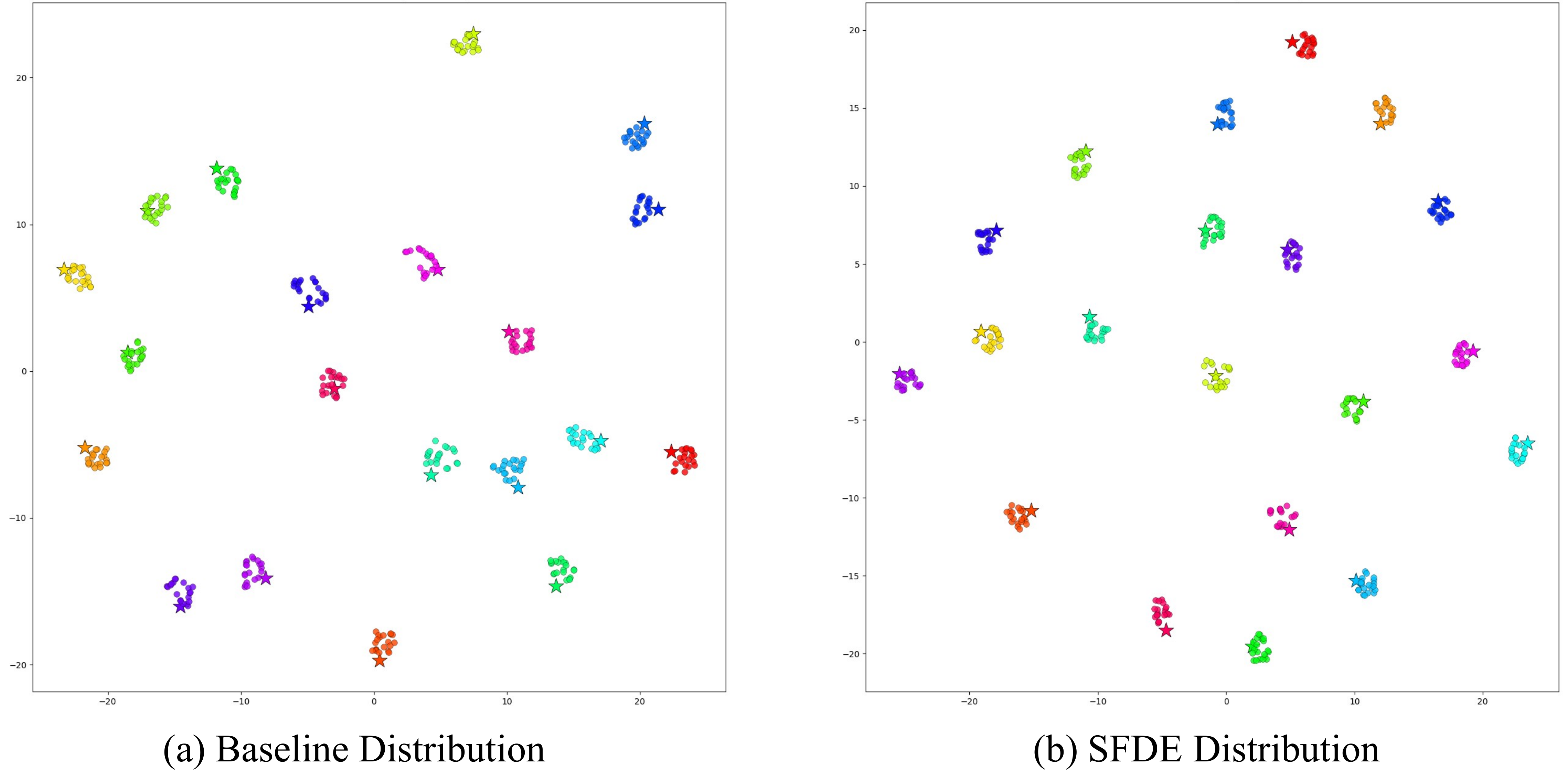}
    \caption{Visualization of feature embeddings in a 2D feature space. We select 40 geo-locations from the test set, samples with the same color correspond to the same location, and the star marker denotes the center of the corresponding location.
}
    \label{fig5}  
\end{figure}
\begin{figure}[ht]
    \centering
    \includegraphics[width=3.5in]{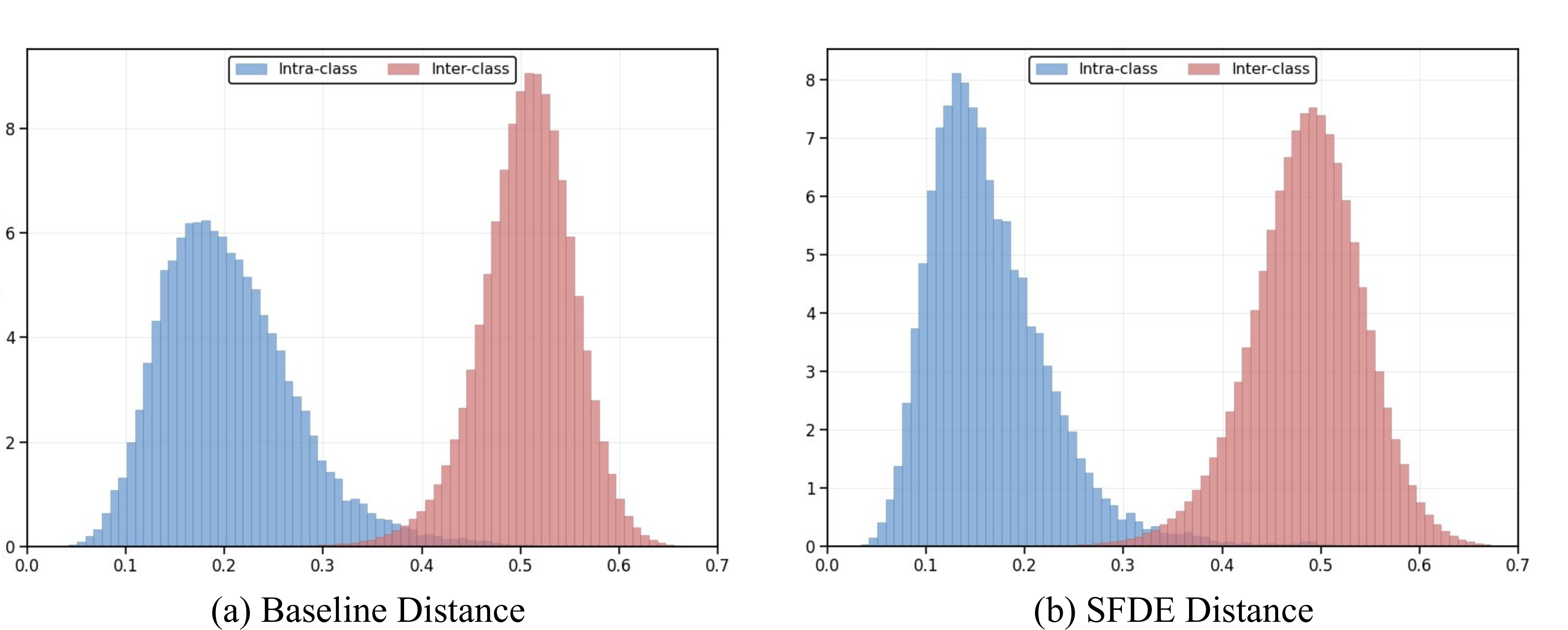}
    \caption{Distance distributions of positive and negative sample pairs in the test set. Blue and red denote the distance distributions of positive (intra-class) and negative (inter-class) sample pairs, respectively.
}
    \label{fig6}  
\end{figure}
\begin{figure}[ht]
  \centering
  \includegraphics[width=3.4in]{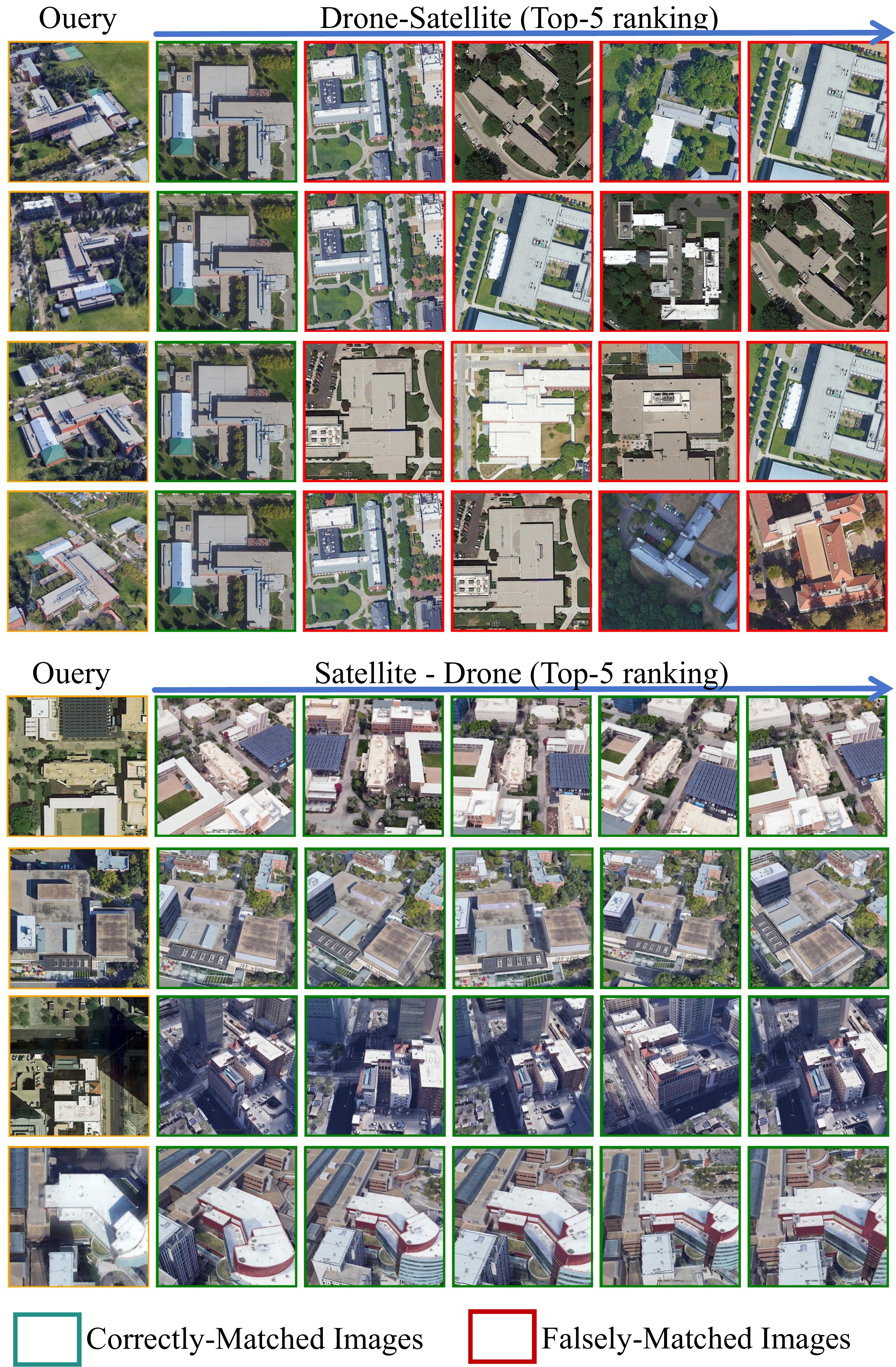}
  \caption{Top-5 retrieval results on the University-1652 dataset. Green bounding boxes indicate correctly matched images, while red bounding boxes denote incorrectly matched images.
}
  \label{fig7}  
\end{figure}
\begin{figure}[ht]
    \centering
    \includegraphics[width=3.5in]{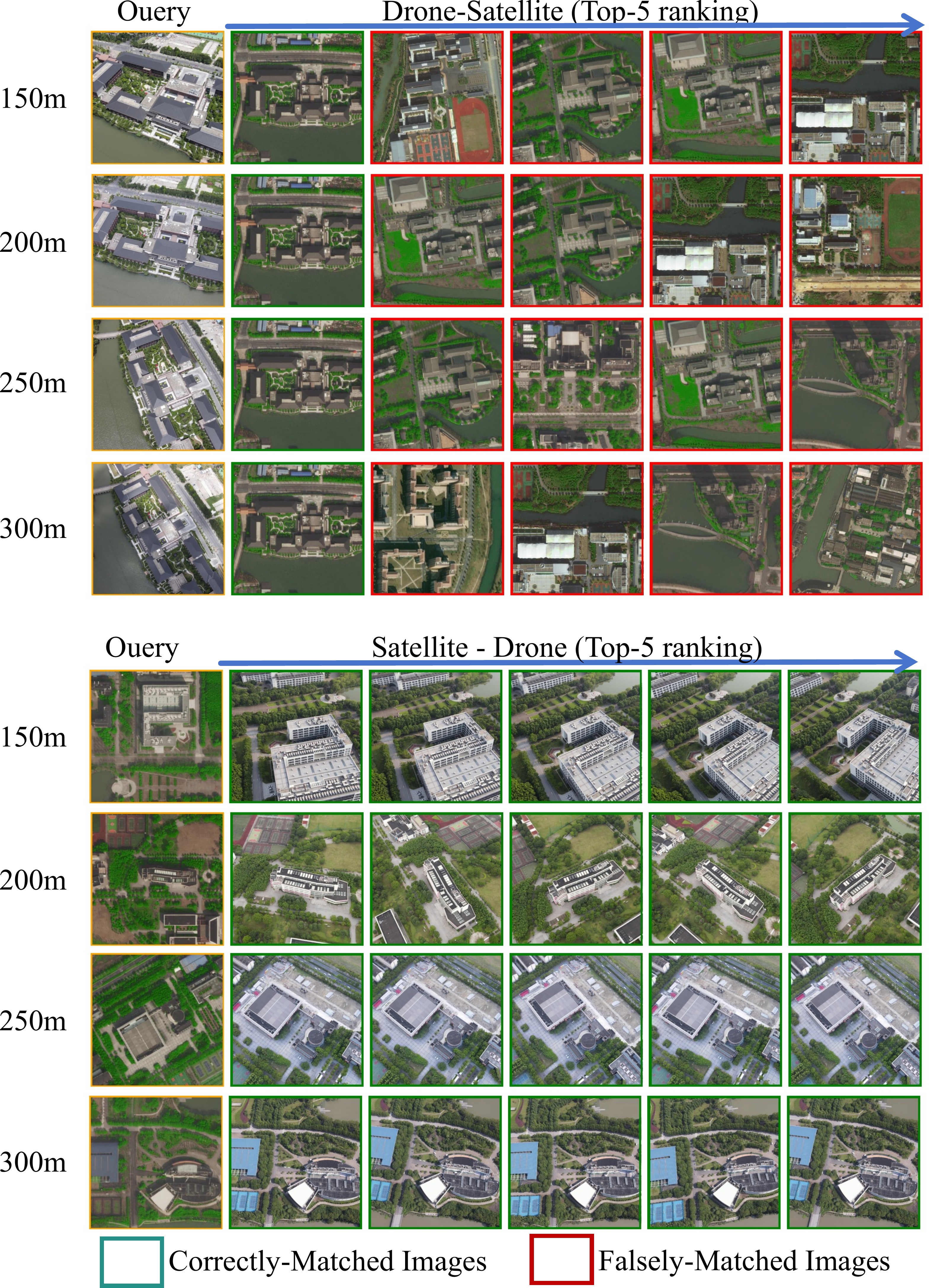}
    \caption{Top-5 retrieval results on the SUES-200 dataset. Green bounding boxes indicate correctly matched images, while red bounding boxes denote incorrectly matched images.
}
    \label{fig8}  
\end{figure}

{\color{black}
Building on the ablation analysis, we further examine the influence of different loss-weight settings on performance through hyperparameter sensitivity experiments involving three key loss terms, namely the global semantic classification loss, the local InfoNCE contrastive loss, and the frequency domain stability alignment loss. As reported in Table~\ref{tab:hp_univ1652_2}, the global semantic loss achieves its optimal performance at a weight of 0.1, under which the Drone$\rightarrow$Satellite task attains an R@1 of 93.75\%. Increasing the weight to 0.2 reduces the R@1 to 92.23\%, and a further increase to 0.5 leads to an additional decline to 90.64\%. In the Satellite$\rightarrow$Drone setting, performance exhibits higher sensitivity to the global loss weight, with the AP decreasing to 88.71\% at a weight of 0.5. This behavior indicates that excessively strong global supervision shifts emphasis away from spatial layout cues that are more critical for cross-view matching.

The influence of the local contrastive loss is summarized in Table~\ref{tab:hp_univ1652_3}. The best performance is obtained when the loss weight is set to 1.0. Reducing the weight to 0.8 results in a decrease of 1.50 percentage points, as weaker contrastive supervision limits the discriminability of the embedding space. Conversely, increasing the weight to 1.2 causes a decline of 1.11 percentage points, suggesting that overly strong constraints may introduce overfitting. Performance variations on the Satellite$\rightarrow$Drone task remain relatively small, indicating that satellite images rely less on local metric constraints.

The impact of the frequency domain alignment loss is presented in Table~\ref{tab:hp_univ1652}. This loss term reaches its optimal weight at 1.3, which is higher than those of the other two losses and reflects the importance of frequency domain alignment in the overall framework. Decreasing this weight reduces the contribution of phase information, leading to R@1 values of 93.16\% and 92.83\%, respectively. Increasing the weight beyond the optimal point results in overfitting, with R@1 declining to 93.34\% and 93.25\%. Overall, the optimal hyperparameter configuration follows a weight ratio of 1:10:13 across global, local, and frequency components, corresponding to a hierarchical optimization strategy in which frequency domain stability plays a primary role, local geometric consistency provides auxiliary support, and global semantic supervision offers complementary guidance.

}

\subsection{Feature Distribution}

To quantitatively assess the feature representation behavior of SFDE, we perform a visualization analysis of intra-class and inter-class distance distributions. We visualize the distance statistics for the baseline method and SFDE in Figs.~\ref{fig5} and~\ref{fig6}. For the baseline, the intra-class and inter-class distance distributions exhibit substantial overlap, which suggests limited separability between positive and negative sample pairs in the feature space. By comparison, SFDE produces a more structured distance distribution, where the intra-class distances concentrate toward lower values and the inter-class distances shift toward higher values, reflecting reduced intra-class dispersion and improved inter-class separation.

In the two dimensional projection space, the baseline method shows a scattered arrangement in which samples from the same class fail to form compact groups, and the boundaries between different classes remain indistinct. By contrast, the feature space produced by SFDE displays a more organized clustering pattern, in which cross-view samples from the same class aggregate within localized neighborhoods to form compact clusters, while different classes remain separated by sufficient distances, leading to clearer inter-class boundaries.

\subsection{Retrieval Results}

To further examine the practical effectiveness of SFDE for cross-view matching, we present a visualization analysis of retrieval results on the University-1652 dataset. Representative retrieval examples on University-1652 are illustrated in Fig.~\ref{fig7}, where green bounding boxes indicate correct matches and red bounding boxes denote mismatches. In the bidirectional retrieval settings, SFDE exhibits stable matching behavior across different query and gallery configurations. Additionally, in SUES-200, as shown in Fig~\ref{fig8}, SFDE consistently maintains accurate matching results across different flight altitudes.

\section{Discussion}

SFDE distinguishes itself through coordinated integration of spatial and frequency domain features. Unlike prior methods that predominantly rely on local spatial alignment or shallow frequency enhancement~\cite{chen2025multilevel}, SFDE adopts a three branch parallel architecture that jointly models global semantic consistency, local geometric sensitivity, and frequency stability alignment. This design allows spatial and frequency representations to complement each other and mitigates the limitations inherent in single-domain feature modeling. Experimental results demonstrate that even with a lightweight configuration, SFDE achieves strong retrieval performance on University-1652, SUES-200, and Multi-weather University-1652 while maintaining stable behavior across flight altitude variations and adverse weather conditions. These results indicate that frequency domain stability contributes to robustness when spatial features are degraded by geometric perturbations. Despite these strengths, several limitations remain. The frequency domain branch relies on offline Fourier transforms, potentially limiting efficiency for ultrahigh-resolution images. Future work will investigate differentiable wavelet transforms or approximate frequency domain operations to further improve computational efficiency. In addition, although the multiobjective loss enables joint optimization of the three branches, inter-branch interaction is achieved implicitly through backpropagation. Exploring explicit cross branch interaction mechanisms, such as attention based fusion or feature distillation, represents another promising direction.

\section{CONCLUSION}

This paper addresses CVGL between UAV and satellite images by proposing the Spatial and Frequency Domain Enhancement Network (SFDE) to mitigate cross-domain feature discrepancies. By integrating representations from the spatial and frequency domains within a unified parallel architecture, SFDE enhances feature stability and improves robustness against geometric perturbations. Extensive experiments across multiple benchmarks validate the effectiveness of the proposed approach. SFDE achieves strong retrieval performance on University-1652, SUES-200, and Multi-weather University-1652, while maintaining stable behavior under variations in flight altitude and adverse weather conditions. Ablation results further highlight the critical contribution of stability alignment in the frequency domain to the joint optimization process. These results demonstrate that SFDE provides a robust and efficient solution for CVGL under diverse conditions, thereby supporting its practical applicability in GNSS-denied environments.

\vspace{8pt}
\noindent \textbf{CRediT authorship contribution statement}

Hongying Zhang: Conceptualization; Methodology; Writing -- original draft; Supervision; Funding acquisition. Shuaishuai Ma: Investigation; Formal analysis. Hongying Zhang and Shuaishuai Ma: Writing -- review \& editing.

\vspace{8pt}
\noindent \textbf{Declaration of competing interest}

We declare that we do not have any commercial or associative
interest that represents a conflict of interest in connection with the
work submitted.

\vspace{8pt}
\noindent \textbf{Acknowledgments}

This work was supported by the Graduate Research Innovation Grant Program of Civil Aviation University of China (YJSKC05005)

\footnotesize
\setlength{\bibsep}{0.0pt}
\bibliographystyle{elsarticle-num}
\bibliography{reference}

\end{document}